%% file: main.tex
\documentclass[letterpaper]{article} 
\usepackage{aaai25}  
\usepackage{times}  
\usepackage{helvet}  
\usepackage{courier}  
\usepackage[hyphens]{url}  
\usepackage{graphicx} 
\def\UrlFont{\rm}  
\usepackage{natbib}  
\usepackage{caption} 
\usepackage{amsmath,amssymb,amsfonts}
\usepackage{pgfplots}
\usetikzlibrary{pgfplots.statistics, pgfplots.colorbrewer, pgfplots.groupplots, shapes.geometric, decorations.markings, arrows, decorations.text, backgrounds,fit,matrix}
\usepgfplotslibrary{fillbetween}
\usepackage{pgfplotstable}
\pgfplotsset{compat=1.15}
\usetikzlibrary{external}
\usetikzlibrary{positioning, calc, fit}
\usepackage{listofitems} 
\usepackage[ruled,vlined,linesnumbered]{algorithm2e}
\SetKwInput{KwInput}{Input}
\SetKwInput{KwOutput}{Output}
\SetKwFunction{FMain}{ReachAnalysis}
\SetKwFunction{FBackMain}{BackReachAnalysis}
\SetKwFunction{FReach}{ComputeReach}
\SetKwFunction{FBackReach}{ComputeBackReach}
\SetKwFunction{FReachCell}{ComputeReachCells}
\SetKwFunction{FBackReachCell}{ComputeBackReachCells}
\SetKwProg{Fn}{Function}{:}{}
\usepackage{siunitx}
\usepackage[super]{nth}
\usepackage{caption,subcaption,booktabs}
\usepackage{multirow}
\usepackage{enumitem}
\usepackage{hhline}

\title{Scalable Surrogate Verification of Image-based Neural Network Control Systems Using Composition and Unrolling}
\author{
    Feiyang Cai\textsuperscript{\rm 1,3},
    Chuchu Fan\textsuperscript{\rm 2},
    Stanley Bak\textsuperscript{\rm 1}
}
\affiliations{
    \textsuperscript{\rm 1}Department of Computer Science, Stony Brook University\\
    \textsuperscript{\rm 2}Department of Aeronautics and Astronautics, Massachusetts Institute of Technology\\
    \textsuperscript{\rm 3}School of Computing, Clemson University\\

    feiyang@clemson.edu, chuchu@mit.edu, stanley.bak@stonybrook.edu
}

\usepackage{bibentry}

\begin{document}

\maketitle

\input{abstract}

\input{introduction}
\input{background}
\input{methodology}

\input{experiments}

\input{conclusion}

\clearpage
\input{acknowledgments}
\bibliography{aaai25}


\clearpage
\appendix
\renewcommand{\thesection}{\Alph{section}.\arabic{section}}
\setcounter{section}{0}
\setcounter{table}{0}
\setcounter{figure}{0}
\renewcommand{\thetable}{S\arabic{table}}
\renewcommand{\thefigure}{S\arabic{figure}}

\input{appendix}

%

\end{document}

%% file: abstract.tex
\begin{abstract}

	Verifying safety of neural network control systems that use images as input is a difficult problem because, from a given system state, there is no known way to mathematically model what images are possible in the real-world. 
	We build upon recent work that considers a surrogate verification approach,  training a conditional generative adversarial network (cGAN) as an image generator in place of the real world.
	This setup enables set-based formal analysis of the closed-loop system, providing analysis beyond simulation and testing. 
	While existing work is effective on small examples, excessive overapproximation both within a single control period (one-step error) and across multiple periods (multi-step error) limits its scalability. 
	We propose approaches to overcome these errors. 
	First, we address one-step error by composing the system's dynamics along with the cGAN and neural network controller, without losing the dependencies between input states and the control outputs as in the monotonic analysis of the system dynamics.
	%
	Second, we reduce multi-step error by repeating the single-step composition,
	essentially unrolling multiple steps of the control loop into a large neural network. 
	We then leverage existing network verification algorithms to compute accurate reachable sets for multiple steps, avoiding the accumulation of abstraction error at each step.
	We demonstrate the effectiveness of our approach in terms of both accuracy and scalability using 
	two case studies.
	%
	On the aircraft taxiing system, 
	the converged reachable set is  $175\%$ larger using the prior baseline method compared with our proposed approach.
	On the emergency braking system, with 24$\times$ the number of image output variables from the cGAN, the baseline method fails to prove any states are safe, whereas our improvements enable set-based safety analysis.
\end{abstract}

%% file: introduction.tex
\section{Introduction}

Neural networks are key enablers of image-based control systems, where applications span from autonomous vehicles~\cite{chen2015deepdriving} to industrial robotics~\cite{levine2018learning}.
Unfortunately, neural networks rarely come with guarantees of robustness or worst-case behaviors 
due to the inherent non-zero error rates and are often
susceptible to adversarial attacks~\cite{boloor2020attacking,cai2020detecting}.
%
In safety-critical scenarios, ensuring the safety and reliability of neural network control systems (NNCS) is paramount. 
The deployment of such systems demands rigorous \emph{closed-loop} verification to show that hazardous situations are avoided.
Although considerable strides have been made in verifying NNCS~\cite{sun2019formal,schilling2022verification},
verification of \emph{image-based} NNCS has recently gained attention~\cite{hsieh2022verifying,astorga2023perception,puasuareanu2023closed,arjomandbigdeli2024verification}.
This task is more difficult---image observations are high-dimensional inputs and vision networks are typically more complex than control networks, employing convolutional or transformer layers.
Scalability aside, such verification problems are even hard to formulate, since reasoning over the set of images possible in the real world is not a precise mathematical statement.
Recent approaches have used geometric camera models to accurately represent the perception process, but they are not applicable to complex scenarios~\cite{habeeb2023verification}.

One approach to this problem is using generative models to replace the perception system for the verification of image-based NNCS~\cite{katz2022verification}. 
This first trains a conditional generative adversarial network (cGAN)~\cite{DBLP:journals/corr/MirzaO14} to approximate the perception system, generating images based on system states.
The concatenation of cGAN and controller results in a unified network controller with low-dimensional state inputs. This surrogate controller allows existing verification methods~\cite{xiang2018reachability}, which combine neural network verification tools and reachability analysis, to be applied.

However, the previous verification methodology~\cite{katz2022verification} suffers from being overly conservative, computing the reachable sets with significant overapproximation, caused by two main sources: \textit{one-step error}
and \textit{multi-step error}.
%
Within a single control period, the prior work
computes the intervals of the control outputs and then applies monotonic analysis of the system dynamics to obtain the reachable sets, 
ignoring
the dependencies between the input states and control outputs, causing one-step error.
Additionally, discrete abstraction of reachable sets after each step introduces multi-step error. The interaction of these two sources of error produces considerable overapproximation, potentially leading to false positives.
As image-based NNCS grow in complexity and incorporate modern neural network architectures, extending the verification techniques to encompass these complexities presents additional challenges.

This work builds upon the surrogate verification approach using cGAN and focuses on mitigating overapproaximation within the existing method.
The first improvement is to compose the discrete-time dynamics along with the cGAN and neural network controller. This composition preserves the 
state-control dependencies, 
reducing one-step error.
We introduce two distinct composition options to accommodate varying network scales and system dynamics.
One option is incorporating the dynamics as an additional layer within the networks, allowing direct handling by neural network verification tools.
This approach may encounter challenges when nonlinear dynamics are not supported by these tools 
or when overapproximation of nonlinear functions in the dynamics causes infinite outputs.
The other option combines neural network exact analysis~\cite{bak2021nfm} with the reachability algorithms from hybrid systems~\cite{althoff2008reachability}.
This method computes more accurate reachable sets using the star set representation~\cite{duggirala2016parsimonious}, further reducing the one-step error.
This approach is applicable when the network's scale is moderate and the network exclusively employs $\mathsf{ReLU}$ activations.
Instead of limiting the verification scope to a single step,
we propose another improvement by unrolling multiple steps of the control loop into a unified, larger neural network.
We then leverage state-of-the-art network verification tools, specifically $\alpha$-$\beta$-CROWN~\cite{zhang2018efficient} and nnenum~\cite{bak2021nfm}, to compute reachable sets over multiple steps. This unrolling strategy reduces the frequency of abstraction operations and, consequently, minimizes the multi-step error.

We evaluate our method with two case studies.
The first focuses on an autonomous aircraft taxiing system~\cite{katz2022verification}, 
featuring a feed-forward neural network controller. 
The evaluation demonstrates that the proposed approach significantly reduces overapproximation, and the reachable set upon convergence using the prior baseline method is $175\%$ larger than with our proposed approach.
Therefore, our method can verify scenarios where the existing approach fails.
The second case is an advanced emergency braking system~\cite{cai2020iccps}, 
incorporating modern image-based controllers in the loop.
This work stands as the first attempt to conduct reachability analysis for a closed-loop system with convolutional and transformer layers.

%% file: background.tex
\section{Background}
This section begins by presenting the system model and formulating the verification problem for image-based NNCS. We then introduce the surrogate approximation of the perception system using generative models. 
Finally, we discuss the challenges 
of verifying such surrogate systems.


\subsection{Problem Formulation}
\begin{figure*}[!t]
	\centering
	\includegraphics[width=0.65\textwidth]{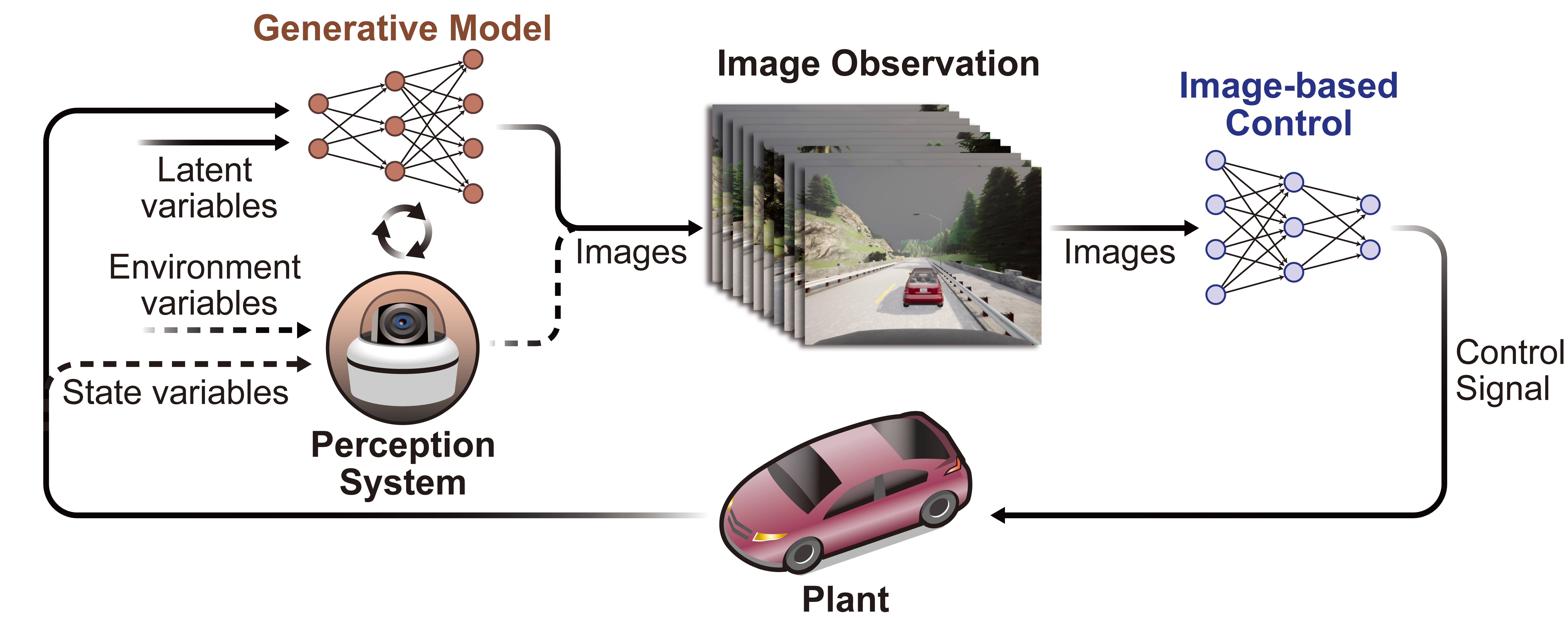}
	\caption{Simplified architecture of an image-based neural network control system and its surrogate system.} 
\label{fig:arch}
\end{figure*}

\newtheorem{definition}{Definition}
\newtheorem{problem}{Problem}
\begin{definition}[System Model]
	Consider an autonomous system with an image-based neural network controller, as depicted in Fig.~\ref{fig:arch}.
	At step $k$, the ego system is in state $x_k \in \mathcal{X}$. 
	The \textit{perception system} 
	$\mathsf{P}$, employs a camera sensor to observe the surrounding environment $e_k \in \mathcal{E}$, and
	translates these observations into a representative image  $o_k \in \mathcal{O}$.
	The image $o_k$
	is then input into an \textit{image-based controller} $\mathsf{C}$ 
	to generate a control command $u_k \in \mathcal{U}$ to accomplish a specific task.
	Driven by $u_k$, the system transitions from state $x_k$ to a subsequent state $x_{k+1}$ 
	according to discrete \textit{dynamics} $\mathsf{D}$, thereby closing the loop.
	The system's evolution over a single step 
	depends on both the current state $x_k$ and the environment status $e_k$, and is expressed as:
	\begin{equation}
		x_{k+1} = \mathsf{D}(x_k, \mathsf{C}(\mathsf{P}(x_k, e_k))).
		\label{eq:step_atom}
	\end{equation}
\end{definition}



\begin{definition}[System Evolution]
	\label{def:system_evolution}
	The system starts from an 
	initial state $x_0$ within a set $\mathcal{I} \subseteq\mathcal{X}$.
	At each time step $i$ between $0$ and $k-1$,
	the corresponding environment is represented as $e_i\subseteq\mathcal{E}$. 
	Consequently, the system state $x_k$ evolves from $x_0$ 
	by unrolling the dynamics defined in Eq.~(\ref{eq:step_atom}). 
	This system evolution depends on the initial state $x_0$ and the series of the environment parameters 
	$e_0, \ldots, e_{k-1}$ over time:
	\begin{equation}
		x_k = \mathsf{System} (x_0, \{e_0, \ldots, e_i, \ldots, e_{k-1}\}).
		\label{eq:system_evolution}
	\end{equation}
\end{definition}

\begin{definition}[Reachable Set]
	The reachable set at step $k$ is
	\begin{equation*}
		\begin{split}
			R_k = & \{\mathsf{System} (x_0, \{e_0, \ldots, e_{k-1}\})\\ 
			&| ~ \forall x_0  \in \mathcal{I}, ~ \forall_{i\in [0, {k-1}]}\  e_i  \in \mathcal{E} \}.	
		\end{split}
		\label{eq:reachable_set}
	\end{equation*}
	Here, $R_0$ denotes the reachable set at time step $0$, which is equal to the initial set $\mathcal{I}$.
	Furthermore, the reachable set over the time interval $[0, k]$ is defined as
	$R_{[0, k]} = \bigcup\nolimits_{i\in [0, k]} R_i$.
%
\end{definition}

\begin{problem}[Safety Verification for Image-Based NNCS]
	\label{problem:verification_actual_system}
	The objective of the safety verification for image-based NNCS is to 
	check whether the computed reachable set adheres to the system's safety property. 
	Formally, given an unsafe region $\mathcal{U} \subseteq \mathcal{X}$, 
	the safety of the NNCS within a bounded time step $k_{\text{max}}$
	can be verified if and only if the following condition is satisfied, that is $R_{[0, k_{\text{max}}]} \cap \mathcal{U} = \emptyset$.
	
\end{problem}

\subsection{Perception System Approximation}
\label{sec:abstraction_cgan}
One of the most significant challenges in verifying image-based 
NNCS
is formalization of the perception system. 
Consider a camera used for autonomous driving. 
Formalizing a specification like ``all images with a car that is $5$ meters ahead are predicted as such"
is difficult due to the interdependence of the captured images on multiple factors,
including both the system state $x$ (e.g., distance between the ego car and the leading car) and environment $e$ (e.g., appearance of the car, weather condition, and other objects on the road).  
%
The challenge lies in providing a clear mathematical and comprehensive description for such a specification.
%

	An alternative approach \cite{katz2022verification} involves using images generated by a cGAN to replace those rendered by the actual perception system for verifying the image-based NNCS. In this context, the specification can be formalized as ``all images that can be produced by a specific cGAN are predicted to be a car $5$ meters ahead."
	Such a specification is practical to work with, and in this paper, we embrace this methodology to establish the verification problem for the closed-loop system. 
	Of course, safety of the cGAN does not guarantee safety in the real world, but this approach does offer an analysis method for this class of systems beyond just simulations and tests.
	%
	%
	The cGAN learns the distribution of images conditioning on auxiliary information and can be used to approximate 
	the original perception system $\mathsf{P}$ into a \textit{surrogate perception system} $\hat{\mathsf{P}}$. 
	The generator of the cGAN maps the conditional information $c$ and a set of latent variables $z$ into an image observation $\hat{o}$.
	When the cGAN serves as a perception surrogate, the conditional information $c$ corresponds to the system states $x$, guiding the generator to create images relevant to the desired information. 
	%
	%
	The environmental variables $e$ are left uncontrolled and represented by latent variables $z$ within the cGAN. 
	
	\begin{definition}[Surrogate System Model]
		The one-step system evolution with the surrogate perception system 
		is $x_{k+1} =  \mathsf{D}(x_k, \mathsf{C}(\hat{\mathsf{P}}(x_k, e_k))).$
	\end{definition}
	
	\begin{problem}[Verification for Surrogate System]
		\label{problem:verification_surrogate_system}
		Instead of verifying the actual system as defined in Problem~\ref{problem:verification_actual_system},
		this paper focuses on the verification of the surrogate system.
	\end{problem}
	
	\subsection{Verification Challenges}
	The concatenation of the cGAN and the image-based controller into a unified neural network transforms the entire system into a state-based NNCS.
	In this way, the image-based NNCS can be analyzed using existing methodologies designed for verifying state-based NNCS~\cite{xiang2018reachability}.
	However, computing reachable sets for an NNCS can still be challenging, especially given the complexity of the network performing both image generation and processing tasks.
	%
	%
	In contrast, in the most recent NNCS verification competition, ARCH-COMP 2023~\cite{lopez2023arch}, the analyzed neural networks generally had only a few dozen neurons and up to $6$ layers.
	Most conventional NNCS verification tools~\cite{bogomolov2019juliareach}, including those participating the competition, 
	lack support for convolutional networks. Even those that do offer support~\cite{tran2020nnv},
	encountering large-scale networks leads to significant overapproximations at each step. 
	These overappproximations accumulate and expand, causing the tool to fail.
	
	%
	
	In earlier work, to practically address the verification of the surrogate image-based NNCS, input images are downsized to small grayscale images, and the cGAN and controller are implemented using feed-forward neural networks. 
	Still, 
	this method exhibits large overapproximation of the reachable sets.
	Such substantial overapproximation could result in instances where the overapproximated reachable sets intersect with unsafe regions, causing safety verification to fail.
	%
	As contemporary image-based control systems grow in complexity, with high-resolution images and advanced architectures like convolutional and transformer layers, verification becomes even more challenging.
	%
	%
	\textbf{The main contribution of this paper is to address these challenges and reduce the overapproximation error.}

%% file: methodology.tex
\section{Methodology}
\label{sec:methodology}
This section analyzes the causes of overapproximation in the prior method
and proposes a method to mitigate this issue.
%
%

\subsection{Overapproximation Analysis}
\label{sec:overapproximation_analysis}

Consider Problem~\ref{problem:verification_surrogate_system}, where a surrogate image-based NNCS
starts from an initial set $R_0$. The verification task is to determine whether the reachable set up to time $k_\text{max}$ satisfies the safety property, 
i.e, avoiding any intersection with the unsafe set $\mathcal{U}$. 
%
Neural network verification tools are commonly employed to prove properties over the
network's input and output.
Therefore, if the system's evolution function, as defined by Eq.~(\ref{eq:system_evolution}), is combined into a unified neural network $\mathsf{NN}$, 
the closed-loop system property can be verified using network verification tools. 
Formally, the set of possible outputs of network (system state at step $k_\text{max}$) is:
\begin{equation*}
	\begin{split}
		\mathsf{Range} (\mathsf{NN}, R_0, \mathcal{E}^{k_\text{max}}) = 
		&\{\mathsf{NN}(x_0, \{e_0, \ldots, e_{k_{\text{max}}-1}\}), \\ 
		&|~\forall x_0\in R_0, ~\forall_{i\in [0, {k_\text{max}-1}]}\  e_i  \in \mathcal{E} \}.
	\end{split}
\end{equation*}
The neural network verification problem is to check if 
$\mathsf{Range} (\mathsf{NN}, R_0, \mathcal{E}^{k_\text{max}}) \cap \mathcal{U} = \emptyset$.
%
However, this task presents challenges.
First, system dynamics often involve nonlinear operations that may not be supported by verification tools. 
Second, these tools may have scalability limitations in that large input sets and complex neural networks may present intractable verification problems. 
Unrolling the system evolution function to span from time step $0$ to $k_\text{max}$ replicates a sequence of  cGANs,  image-based controllers, and dynamics layers.
This composed network can be too complex even for state-of-the-art neural network verification tools.

To address these challenges, a practical approach integrating neural network verification tools with reachability methods was proposed in previous works~\cite{xiang2018reachability}, later extended to verify the surrogate image-based NNCS~\cite{katz2022verification}. This method serves as a baseline for comparison in this paper.
%
The prior approach constructs and analyzes a discrete existential abstraction of the system, with transitions defined using one-step reachability.
The process starts by dividing the state space into a finite number of rectangular cells along a grid $\mathcal{H}$. 
The initial states in $R_0$ are then abstracted using a set of rectangular cells referred to as $\mathcal{C}_0$, where $R_0 \subseteq \mathcal{C}_0 \subseteq \mathcal{H}$.
For each cell $c$ within $\mathcal{C}_0$, the interval bounds on the control commands are determined using a neural network verification tool.
These bounds are combined with a monotonic analysis of the system dynamics to result in an interval of possible one-step successors from state cell $c$, called $\mathcal{R}^c_{1}$. 
For example, in the autonomous aircraft taxiing system we will analyze later, the dynamics updating function for the heading angle error $\theta$ is: $\theta_{k+1} = \theta_k + \frac{v}{L}\Delta t \tan \phi_k$ , where $v$, $L$ and $\Delta t$ are constants, $\phi_k$ is the control signal. As the $\tan$ function is monotonically increasing in the specified operating region, monotonic analysis computes the upper (lower) bound of $\theta_{k+1}$ by simultaneously substituting the upper (lower) bounds for both $\theta_k$ and $\phi_k$ into the dynamics function~\cite{katz2022verification}.

%

\begin{figure}[!ht]
	\centering
	\includegraphics[width=1.0 \columnwidth]{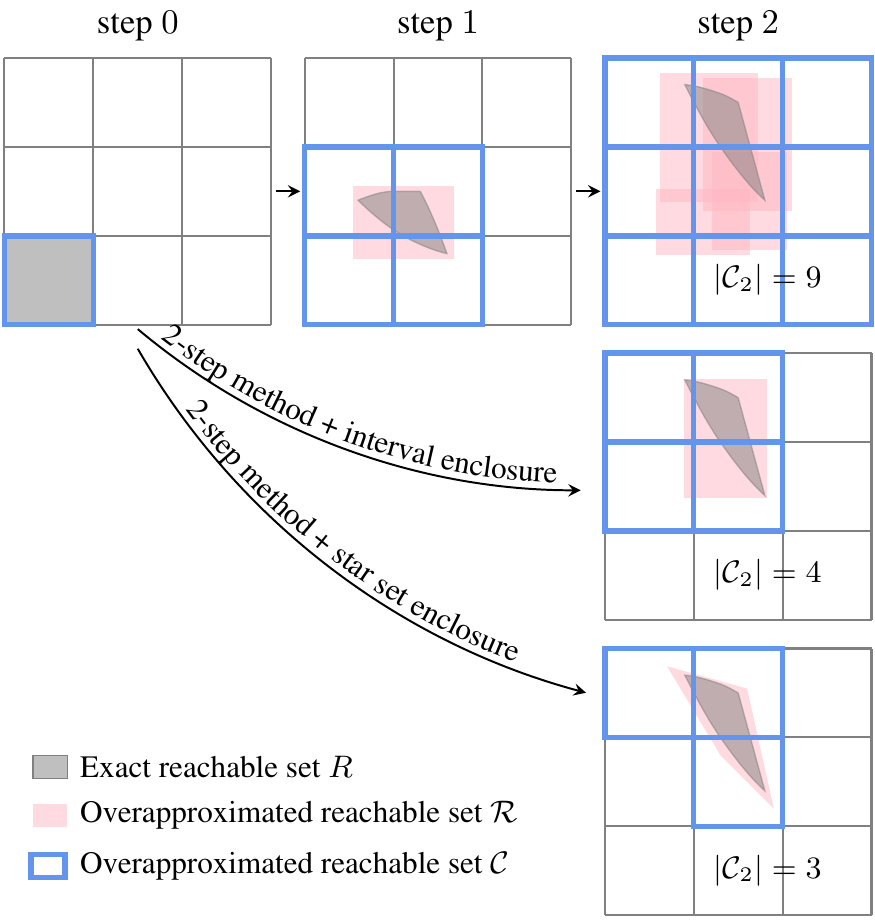}
	\caption{Illustrations of reachable sets using three different methods. 
		In this illustrative example, assume the initial exact reachable set $R_0$ only contains one cell $c$, $R_0 = \mathcal{C}_0 = \{c\}$.}
	\label{fig:method_illustration}
\end{figure}

%
The set of possible successor states from $c$ is then the set of cells $\mathcal{C}^c_{1}$ that overlap with $\mathcal{R}^c_{1}$. 
An overapproximation of the reachable set for the entire initial state is then the union of all successors from individual cells, $\mathcal{C}_{1} = \bigcup_{c\in \mathcal{C}_0} \mathcal{C}^c_{1}$, which overapproximates $\mathcal{R}_1 = \bigcup_{c\in \mathcal{C}_0} \mathcal{R}^c_{1}$, and the exact reachable set $R_1$, denoted as $R_1 \subseteq \mathcal{R}_1 \subseteq \mathcal{C}_1$. This process is repeated iteratively to obtain reachable set $\mathcal{C}_k$ for arbitrary time step $k$. 
The first row of Fig.~\ref{fig:method_illustration_diagram}  illustrates this computation flow from  $R_{0}$ to $\mathcal{C}_{1}$, and the first row of Fig.~\ref{fig:method_illustration} displays the resulting reachable sets from  $R_{0}$ to $\mathcal{C}_{2}$.
Note that the latent space representing the environment is not divided,
and thus, the entire set $\mathcal{E}$ is considered 
when analyzing each cell. 

\begin{figure*}[!t]
	\centering
	\tikzsetnextfilename{method_illustration_diagram}
	\input{figures/illustration_diagram}
	\caption{Computation flowcharts for the baseline (first row), $1$-step (second row), and $2$-step (third row) methods.
	}
	\label{fig:method_illustration_diagram}
\end{figure*}

However, this baseline method introduces considerable overapproximation error, which we call \textit{one-step error} and \textit{multi-step error}.
The one-step error reflects the discrepancy between the overapproximated reachable set $\mathcal{R}$ and the exact reachable set $R$ within a single step. This error arises from two factors within the baseline method. First, the method computes the interval of the control outputs without accounting for the input-output (state-control) dependencies;
second, the method uses monotonic analysis for system dynamics, resulting in an interval enclosure of the reachable set for the next state, 
which is a coarse overapproximation of the exact reachable set.
The one-step error is illustrated in the middle of the first row in Fig.~\ref{fig:method_illustration}, where,
at step $1$, the exact reachable set is a gray curvilinear triangle, while the baseline algorithm overapproximates it as a pink rectangle. 
%
The multi-step error, on the other hand, arises from the abstraction error when transitioning from $\mathcal{R}$ to $\mathcal{C}$. 
As shown in the top middle in Fig.~\ref{fig:method_illustration}, 
after overapproximating the reachable set with the pink rectangle, the baseline method further abstracts it with $4$ blue boxes, introducing the  multi-step error. 
As steps increase, both of errors interact and accumulate, compromising the verification accuracy.
%
%
At time step $2$ (top right, Fig.~\ref{fig:method_illustration}), with the baseline method, all $9$ cells in the graph are considered reachable, while the exact reachable set occupies a curvilinear triangle smaller than $3$ cells.

\subsection{Proposed Improvements}

\subsubsection{Composition}
\label{sec:composition}
%
%
Composing the system dynamics with the cGAN and controller within a single control period can preserve the dependencies  between input states and control outputs, thereby reducing the one-step error.
This computation flow is depicted in the second row of Fig.~\ref{fig:method_illustration_diagram}.

The first proposed method for composition is to directly append the system dynamics 
to the neural networks as an additional layer.   
%
%
Verification tools can then assess the output specifications of a complete step given the input sets, internally preserving the state-control dependencies. 
Rather than computing interval bounds on the control outputs, the tools would be used to compute interval bounds on the successor states.
However, there are drawbacks to this composition method. 
First, it assumes that the system dynamics are either linear or consist only of nonlinear operations supported by a neural network verification tool.
Even if supported, verification tools may still fail to produce bounds.
For instance, many state-of-the-art neural network verification tools, like $\alpha, \beta$-CROWN~\cite{zhang2018efficient},  are based on bounds propagation and refinement.
Nonlinear functions like $\tan$, 
while supported, might have infinite outputs when their ranges are overapproximated---if the input to $\tan$ is initially overapproximated beyond $(-\frac{\pi}{2}, \frac{\pi}{2})$.
In addition, although this composition can result in a tighter interval enclosure of the reachable set compared to the baseline method, it is still not exact and can cause some one-step error.

A second composition approach involves combining dependency-preserving exact neural network analysis methods with reachability analysis algorithms from hybrid systems. 
Unlike bounds propagation methods,
dependency-preserving methods such as those used in nnenum~\cite{bak2021nfm} or NNV~\cite{tran2020nnv} provide the set of possible outputs along with their relationship to inputs, represented as a union of star sets~\cite{duggirala2016parsimonious}. 
(sometimes called constrained zonotopes~\cite{scott2016constrained}).
%
%
%
When dynamics are nonlinear, reachability analysis algorithm, 
specifically the conservative linearization approach~\cite{althoff2008reachability}, is used.
This method soundly overapproximates nonlinear systems as locally linear with added noise to account for the maximum linearization error.
While this approach introduces some overapproximation from the linearization error, 
the reachable set is significantly more accurate than the monotonic analysis approach in the prior work, reducing one-step error.
Additionally, this composition method does not have the infinite output problem when the dynamics involve $\tan$ or other reciprocal functions.
This composition method does have scalability limitations related to the size and type of the neural network, as exact analysis
of the network is required.



\subsubsection{Unrolling}
Multi-step error is attributed to the abstraction operation from reachable set $\mathcal{R}$ to a set of cells $\mathcal{C}$ at each step.
To reduce the frequency of abstractions and, consequently, the multi-step error, 
we propose to unroll multiple control periods and combine them into a single large neural network.
This strategy composes multiple steps of analysis into a single larger operation, to avoid error caused by repeated
abstraction step.
%
The third row of Fig.~\ref{fig:method_illustration_diagram} illustrates this strategy with a $2$-step unrolling.
Increasing the number of unrolling steps $m$ generally leads to a smaller overapproximation error, but a more complex verification problem at each step.
Thus, the selection of $m$ depends on both the system complexity and the verification tool used.

\subsubsection{Improvement Illustration}
The composition and unrolling strategies collectively contribute to reducing overapproximation, as illustrated in the second and third rows of Fig.~\ref{fig:method_illustration}.
The second row shows the method employing
a $2$-step unrolling strategy, 
coupled with the first composition approach wherein the dynamics are integrated as a network layer.
This approach enables direct computation of the interval enclosure for reachable set at step $2$,
which is then abstracted into $4$ cells, as opposed to the $9$ cells in baseline method shown in the first row.
The third row illustrates the method with a $2$-step unrolling strategy, 
while using the composition based on exact neural network analysis and reachability algorithm for the dynamics.
The reachable set at step $2$ is represented by star sets, which aligns more closely with the exact reachable set. By checking the intersection between the star sets and rectangular cells~\cite{wetzlinger2023fully}, the reachable set 
at time step $2$
can be reduced to $3$ cells compared to $4$ cells using the interval enclosure.

\subsubsection{Overall Algorithm}

Here we introduce the overall algorithm for the reachability analysis, with the detailed pseudocode  presented in Algorithm~\ref{alg:reach_analysis},  and a formal justification are provided in Appendix~\ref{appendix:methodology}.
We start by initializing the global reachable set $\mathcal{C}$ with a set of cells $\mathcal{C}_0$ abstracted from the initial set $R_0$.
The abstraction process is formally expressed as $\mathcal{C} = \alpha(\mathcal{R}, \mathcal{H})$, where $\mathcal{H}$ is the set of all rectangular cells in the state space.
Assuming the number of unrolling steps as $m$, 
the reachable set after $m$ steps $\mathcal{R}_{k+m}$ from time step $k$ is denoted as $\mathcal{R}_{k+m} = \FReach_m(\mathcal{C}_{k})$.
It should note that the computation process must be applied for each cell within $\mathcal{C}_{k}$, with results then being combined to obtain the reachable set.
However, using $\FReach_m$ alone allows us to compute reachable sets only when the time steps are multiples of $m$. 
To ensure soundness we must span the entire time domain, requiring $\FReach_1$ through $\FReach_{m-1}$.
Starting from $\mathcal{C}_0$, we compute reachable sets
$\mathcal{R}_{1}$ through $\mathcal{R}_{m}$, which are then abstracted into $\mathcal{C}_{1}$ to $\mathcal{C}_{m}$.
These newly computed sets are accumulated into the global reachable set $\mathcal{C}$.
The reachable set $\mathcal{C}_m$ is used as the starting set for the next iteration.
As $\mathcal{C}_m$ serves an overapproximation of the exact reachable set $R_m$, 
computing the successor reachable sets from $\mathcal{C}_m$ ensures the soundness of the reachability analysis.
The iterative process continues until either the reachable set has converged to an invariant set,
or the time step reaches the bounded step $k_\text{max}$.
Finally, the algorithm returns 
the global reachable set $\mathcal{C}$ 
and a safety flag indicating if $\mathcal{U}$ was reached.
%
We additionally introduce a backward reachability algorithm designed to identify all cells that can be guaranteed to be safe. This algorithm is comprehensively outlined in Appendix \ref{appendix:methodology}. 

\begin{algorithm}[!h]
	\SetAlgoLined
	\DontPrintSemicolon
	\Fn{\FMain}{
		
		\KwInput{$R_0$, initial set}
		\KwInput{$\mathcal{U}$, unsafe region}
		\KwInput{$m$, unrolling steps}
		\KwInput{$k_\text{max}$, termination time step}
		\KwInput{$\alpha$, abstraction function}
		\KwInput{$\mathcal{H}$, rectangular cells defined within the state space}
		
		$\mathcal{C} := \mathcal{C}_0 := \alpha(R_0, \mathcal{H})$\;
		$k:= 0$\;
		$\textit{isConverged}:= \text{false}$\;
		
		\While{$\neg$ isConverged \text{and} $k < k_\text{max}$}  
	{
		\tcc{compute reach set for $k+1$ to $k+m$}
		\For{$i=1$ \KwTo $m$}{
			$\mathcal{R}_{k+i} := \FReach_i(\mathcal{C}_k)$\;
			$\mathcal{C}_{k+i} := \alpha(\mathcal{R}_{k+i}, \mathcal{H})$\;
			$\mathcal{C}:= \mathcal{C} \cup \mathcal{C}_{k+i}$
		}
		$\textit{isConverged}:= (\mathcal{C}_k = \mathcal{C}_{k+m})? \enspace \text{true}:\text{false}$
		
		$k:=k+m$\;
	}
	$\textit{isSafe}:= (\mathcal{C} \cap \mathcal{U} = \emptyset)? \enspace \text{true}:\text{false}$\;
	\KwRet{$\mathcal{C}$, $\textit{isSafe}$}
}
\caption{Proposed reachability algorithm.}
\label{alg:reach_analysis}
\end{algorithm}

%% file: figures/illustration_diagram.tex
\begin{tikzpicture}
	
	\node [draw, rectangle, fill=cyan!20, minimum width=1.5cm, minimum height = 0.8cm, font={\small}, align=center] at (0.4, 0)(abs0){Abstraction};
	\draw [->, line width=0.3mm, >=latex](-1.3, 0)--(abs0) node[above,pos=0.5] {$R_0$};
	
	\node [draw, rectangle, fill=none, minimum width=0.8cm, minimum height = 0.8cm, font=\normalsize, align=center] at (2.6, 0)(perp0){$\hat{\mathsf{P}}$};
	\node [draw, rectangle, fill=none, minimum width=0.8cm, minimum height = 0.8cm, font=\normalsize, align=center] at (4.1, 0)(ctl0){${\mathsf{C}}$};
	\begin{scope}[on background layer]
	\node [draw, fill=yellow!20, dashed, fit=(perp0)(ctl0), minimum height= 1.0cm](comp0){};
	\end{scope}
	\pgfgetlastxy{\firstrectwidth}{\firstrectheight}
	
	\node [draw, rectangle, fill=none, minimum width=0.8cm, minimum height=0.8cm, font=\normalsize, align=center] at (5.6, 0)(dyn0){${\mathsf{D}}$};
	\begin{scope}[on background layer]
		\node [draw, fill=yellow!20, dashed, fit=(dyn0), minimum height= 1.0cm](comp1){};
	\end{scope}
	
	\draw [->, line width=0.3mm, >=latex](abs0)--(comp0) node[above,pos=0.5] {$\mathcal{C}_0$};
	\draw [->, line width=0.3mm, >=latex](perp0)--(ctl0);
	\draw [->, line width=0.3mm, >=latex](comp0)--(comp1);

	\node [draw, rectangle, fill=cyan!20, minimum width=1.5cm, minimum height = 0.8cm, font=\small, align=center] at (7.8, 0)(abs1){Abstraction};
	\draw [->, line width=0.3mm, >=latex](comp1)--(abs1) node[above,pos=0.5] {$\mathcal{R}_1$};
	\draw [->, line width=0.3mm, >=latex](abs1)--(9.5,0) node[above,pos=0.5] {$\mathcal{C}_1$};
	
	
	
	\node [draw, rectangle, fill=cyan!20, minimum width=1.5cm, minimum height = 0.8cm, font=\small, align=center] at (0.4, -1.2)(abs0_1){Abstraction};
	\draw [->, line width=0.3mm, >=latex](-1.3, -1.2)--(abs0_1) node[above,pos=0.5] {$R_0$};
	
	\node [draw, rectangle, fill=none, minimum width=0.8cm, minimum height = 0.8cm, font=\normalsize, align=center] at (2.6, -1.2)(perp0_1){$\hat{\mathsf{P}}$};
	\node [draw, rectangle, fill=none, minimum width=0.8cm, minimum height = 0.8cm, font=\normalsize, align=center] at (4.1, -1.2)(ctl0_1){${\mathsf{C}}$};
	\node [draw, rectangle, fill=none, minimum width=0.8cm, minimum height = 0.8cm, font=\normalsize, align=center] at (5.6, -1.2)(dyn0_1){${\mathsf{D}}$};
	
	\begin{scope}[on background layer]
		\node [draw, fill=yellow!20, dashed, fit=(perp0_1)(ctl0_1)(dyn0_1), minimum height= 1.0cm](comp0_1){};
	\end{scope}
	\draw [->, line width=0.3mm, >=latex](abs0_1)--(comp0_1) node[above,pos=0.5] {$\mathcal{C}_0$};
	\draw [->, line width=0.3mm, >=latex](perp0_1)--(ctl0_1);
	\draw [->, line width=0.3mm, >=latex](ctl0_1)--(dyn0_1);

	\node [draw, rectangle, fill=cyan!20, minimum width=1.5cm, minimum height = 0.8cm, font=\small, align=center] at (7.8, -1.2)(abs1_1){Abstraction};
	\draw [->, line width=0.3mm, >=latex](comp0_1)--(abs1_1) node[above,pos=0.5] {$\mathcal{R}_1$};
	\draw [->, line width=0.3mm, >=latex](abs1_1)--(9.5,-1.2) node[above,pos=0.5] {$\mathcal{C}_1$};

	\node [draw, rectangle, fill=cyan!20, minimum width=1.5cm, minimum height = 0.8cm, font=\small, align=center] at (0.4, -2.4)(abs0_2){Abstraction};

	\draw [->, line width=0.3mm, >=latex](-1.3, -2.4)--(abs0_2) node[above,pos=0.5] {$R_0$};
	
	\node [draw, rectangle, fill=none, minimum width=0.8cm, minimum height = 0.8cm, font=\normalsize, align=center] at (2.6, -2.4)(perp0_2){$\hat{\mathsf{P}}$};
	\node [draw, rectangle, fill=none, minimum width=0.8cm, minimum height = 0.8cm, font=\normalsize, align=center] at (4.1, -2.4)(ctl0_2){${\mathsf{C}}$};
	\node [draw, rectangle, fill=none, minimum width=0.8cm, minimum height = 0.8cm, font=\normalsize, align=center] at (5.6, -2.4)(dyn0_2){${\mathsf{D}}$};
	\node [draw, rectangle, fill=none, minimum width=0.8cm, minimum height = 0.8cm, font=\normalsize, align=center] at (7.1, -2.4)(perp1_2){$\hat{\mathsf{P}}$};
	\node [draw, rectangle, fill=none, minimum width=0.8cm, minimum height = 0.8cm, font=\normalsize, align=center] at (8.6, -2.4)(ctl1_2){${\mathsf{C}}$};
	\node [draw, rectangle, fill=none, minimum width=0.8cm, minimum height = 0.8cm, font=\normalsize, align=center] at (10.1, -2.4)(dyn1_2){${\mathsf{D}}$};
	\begin{scope}[on background layer]	
	\node [draw, fill=yellow!20, dashed, fit=(perp0_2)(ctl0_2)(dyn0_2)(perp1_2)(ctl1_2)(dyn1_2)](comp1_2){};
	\end{scope}	
	\draw [->, line width=0.3mm, >=latex](abs0_2)--(comp1_2) node[above,pos=0.5] {$\mathcal{C}_0$};
	\draw [->, line width=0.3mm, >=latex](perp0_2)--(ctl0_2);
	\draw [->, line width=0.3mm, >=latex](ctl0_2)--(dyn0_2);
	\draw [->, line width=0.3mm, >=latex](dyn0_2)--(perp1_2);
	\draw [->, line width=0.3mm, >=latex](perp1_2)--(ctl1_2);
	\draw [->, line width=0.3mm, >=latex](ctl1_2)--(dyn1_2);
	
	\node [draw, rectangle, fill=cyan!20, minimum width=1.5cm, minimum height = 0.8cm, font=\small, align=center] at (12.3, -2.4)(abs2_2){Abstraction};
	
		\draw [->, line width=0.3mm, >=latex](comp1_2)--(abs2_2) node[above,pos=0.5] {$\mathcal{R}_2$};
		\draw [->, line width=0.3mm, >=latex](abs2_2)--(14.0,-2.4) node[above,pos=0.5] {$\mathcal{C}_2$};
	
\matrix (table) [draw=none, fill=none, below left, matrix of nodes, nodes in empty cells, nodes={inner sep=3.8pt}] 
at (13.6, 0.4) {
	\node [draw, rectangle, dashed, fill=yellow!20, minimum width=0.4cm, minimum height = 0.4cm, label=right:\small Cause one-step error] {}; \\
	\node [draw, rectangle, fill=cyan!20, minimum width=0.4cm, minimum height = 0.4cm, label=right:\small Cause multi-step error] {}; \\
};

\end{tikzpicture}

%% file: experiments.tex
\section{Case Studies}
%
%
%


\input{experiments-autonomous_aircraft_taxi_system}
\input{experiments-advanced_emergency_braking_system}

%% file: experiments-autonomous_aircraft_taxi_system.tex
\subsection{Autonomous Aircraft Taxiing System}
\label{ssec:taxinet}
We use the autonomous aircraft taxiing system, evaluated by the baseline method, to demonstrate our improvements.
%

\subsubsection{System Details}

The system's task is to control the steering of an aircraft moving at a constant speed on a taxiway 
according to nonlinear discrete dynamics: 
$p_{k+1} = p_{k} + v \Delta t \sin \theta_k $ and $\theta_{k+1} = \theta_k + \frac{v}{L} \Delta t \tan \phi_k$.
The aircraft's state is defined by its crosstrack position $p$ and heading angle error $\theta$. 
Here, $v$, $\Delta t$, and $L$ represent the aircraft's taxi speed (\qty[per-mode = symbol]{5}{\meter\per\second}), dynamics updating period (\qty{0.05}{\second}), and the distance between front and back wheels (\qty{5}{\meter}), respectively. 
The control signal $\phi_k$ is generated by an image-based controller running at 1 Hz, which first uses a neural network to predict the state variables $\hat{p}_k$ and $\hat{\theta}_k$, 
followed by a proportional control strategy: $\phi_k = -0.74 \hat{p}_k - 0.44 \hat{\theta}_k$.
%
The perception is approximated using a cGAN conditioned on the aircraft's states $p$ and $\theta$.
Two latent variables from $-0.8$ to $0.8$ are introduced to capture environment variations.
The unified network, which includes both the cGAN and the image-based controller, consists of $8$ fully-connected layers with $\mathsf{ReLU}$ activations. 
Further details on the network architecture and the generated images are available in the Appendix \ref{appendix:aats}.

\subsubsection{Verification Results} 
The state space is defined with $p\in[\qty{-11}{\meter}, \qty{11}{\meter}]$ and $\theta\in[\qty{-30}{\degree}, \qty{30}{\degree}]$.
Consistent with prior work,
we partition the space into a grid of $128 \times 128$ cells with uniform cell width in each dimension. 

We initiate our comparison by contrasting the proposed method with the baseline through a single-cell reachability analysis.
The proposed approach begins by composing the network and the dynamics to preserve the dependencies between input states and control outputs.
We first try to integrate
the system dynamics as an additional layer within the neural networks.
Despite support for $\sin$ and $\tan$ within $\alpha, \beta$-CROWN,
the verification process \emph{fails} in our experiment
due to the infinite output of the $\tan$ function during the initial refinement round, as discussed in Methodology section.
We next try the proposed alternative composition approach, using exact analysis with nnenum and reachability analysis.
%
To account for the nonlinear dynamics,
we employ a conservative linearization technique based on Taylor expansion~\cite{althoff2008reachability}.
Detailed explanations of this technique are provided in
Appendix~\ref{appendix:aats}.

\begin{figure}[!t]
	\centering
	\tikzsetnextfilename{taxi_system_single_cell_propogate}
	\includegraphics[width=1.0\columnwidth]{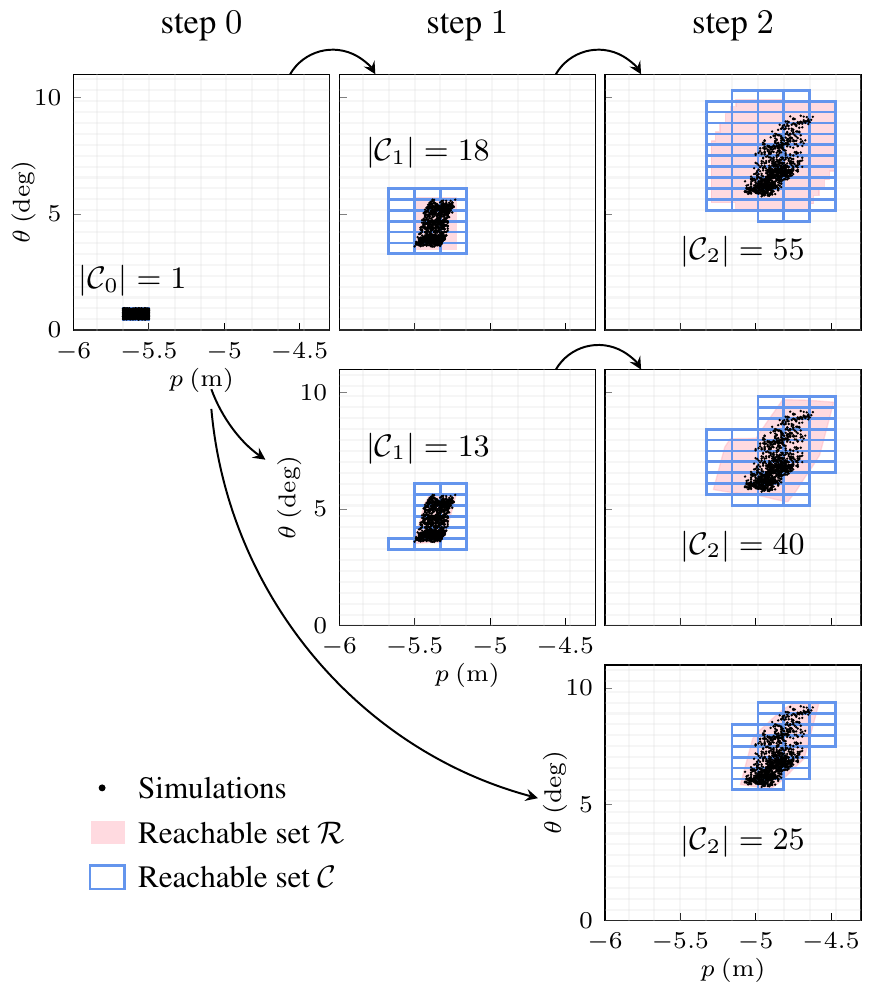}
	\caption{Reachable sets 
		computed using baseline (first row), $1$-step (second row), and $2$-step (third row) methods in the taxiing system. The initial set only contains one cell.}
	\label{fig:taxi_system_single_cell_propogate}
\end{figure}

This reachability method tracks dependencies between the initial state and the next state under a control command, and 
therefore has less overapproximation compared to monotonic analysis, as illustrated in Fig.~\ref{fig:taxi_system_single_cell_propogate}.
Starting from the same single cell (row 1, column 1), the pink reachable set in the proposed method (row 2, column 2), exhibits a smaller area compared to the pink box interval estimated by the baseline (row 1, column 2). 
Furthermore, the number of abstract successor states is reduced from $18$ in the baseline to $13$ when dependency-preserving composition is used.

\begin{figure*}[!ht]
	\centering
	\tikzsetnextfilename{taxi_system_reachable_sets_original_0s}
	\subfloat{\includegraphics{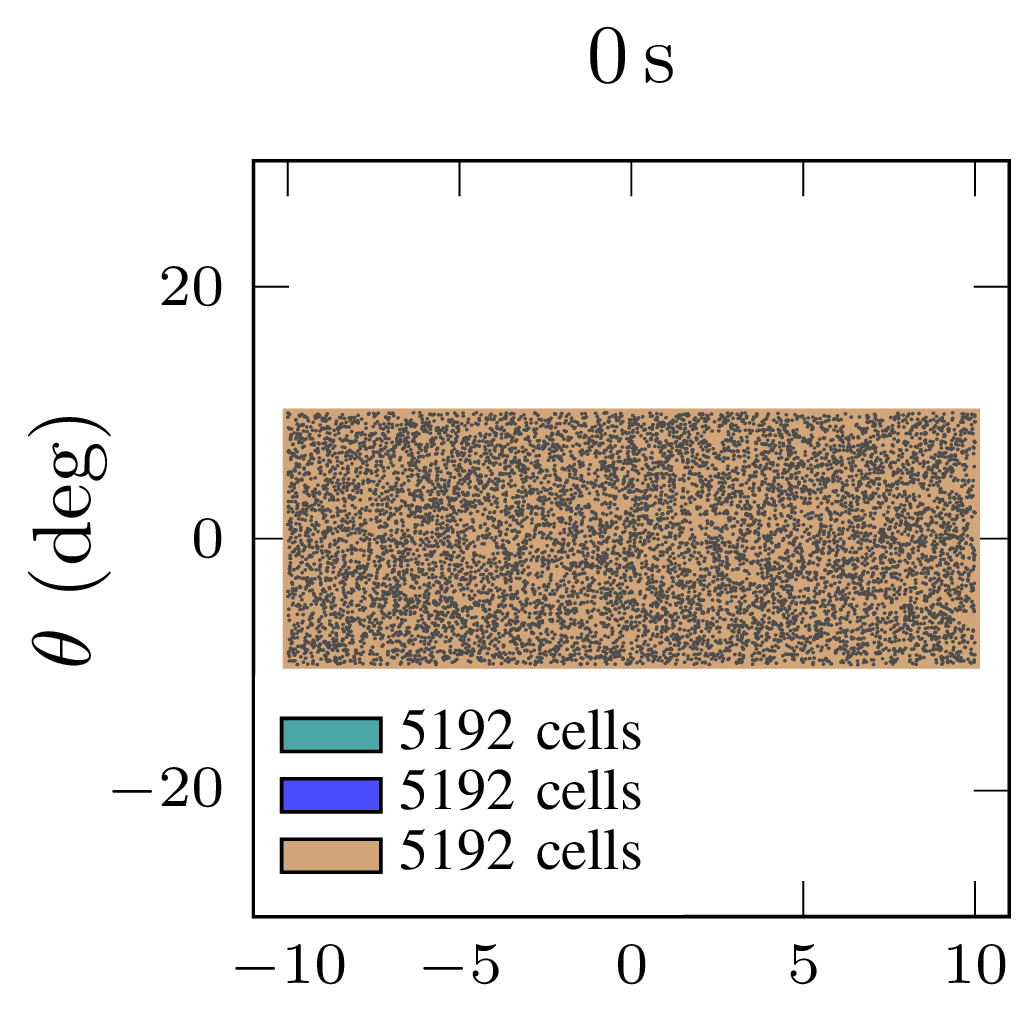}}
	\hspace{-1.1em}
	\subfloat{\includegraphics{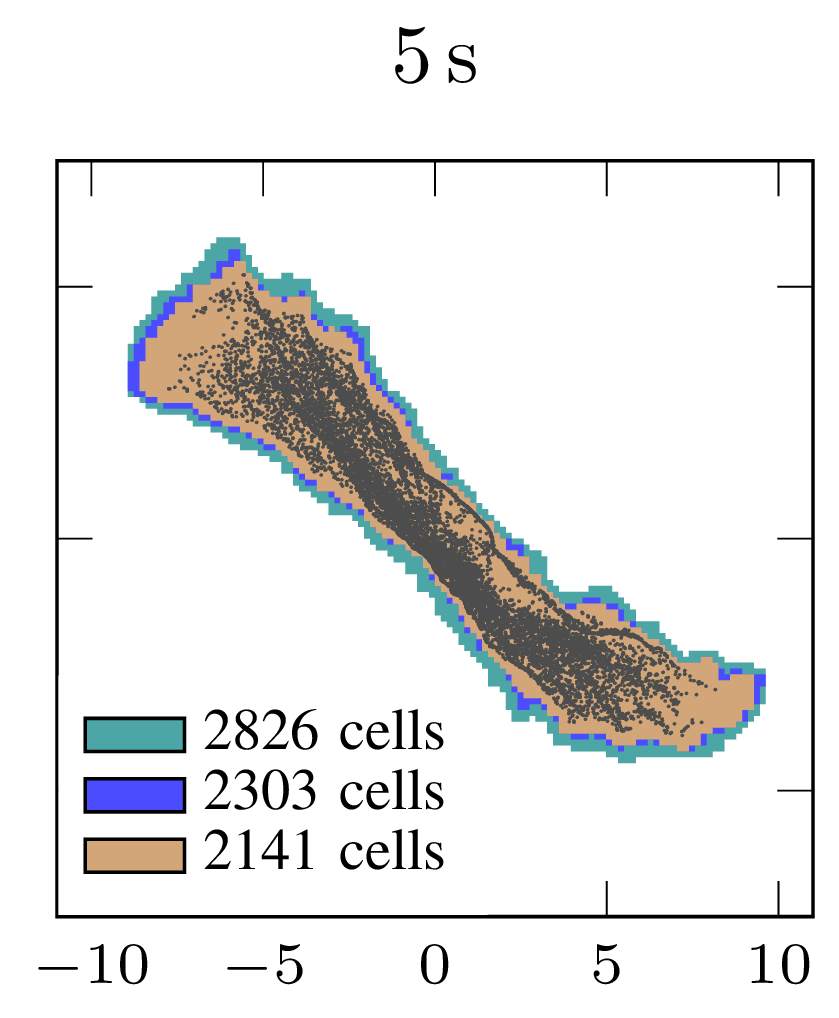}}
	\hspace{-1.1em}
	\subfloat{\includegraphics{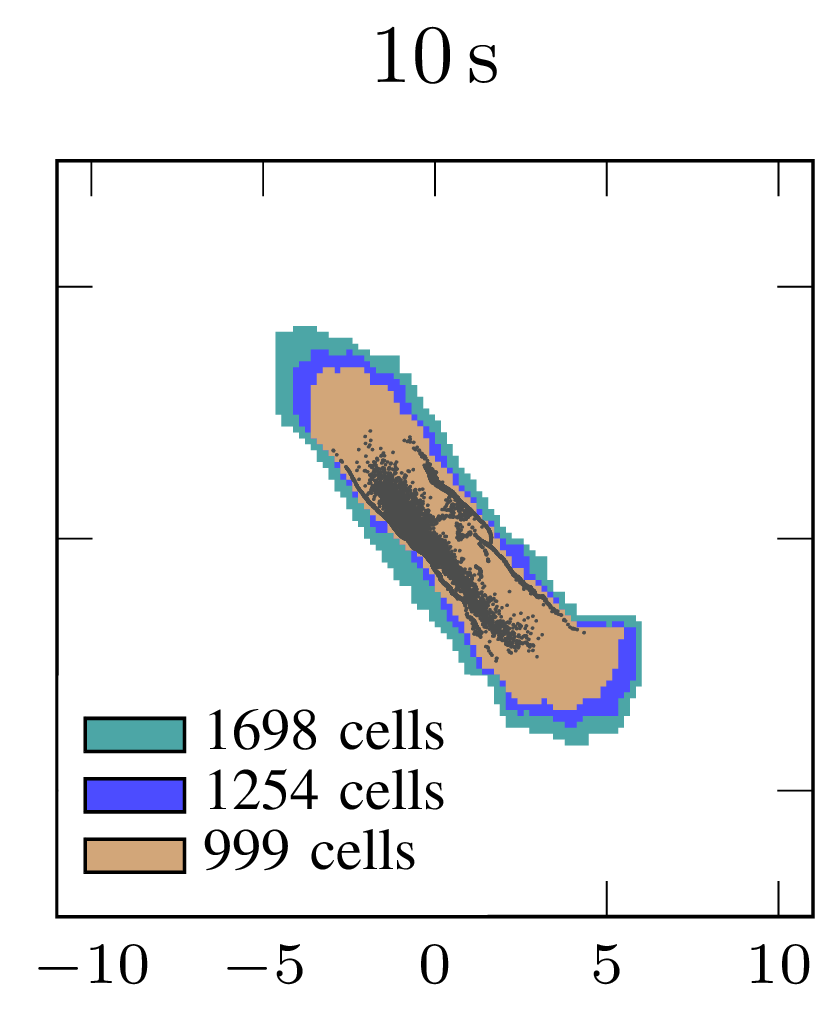}}
	\hspace{-1.1em}
	\subfloat{\includegraphics{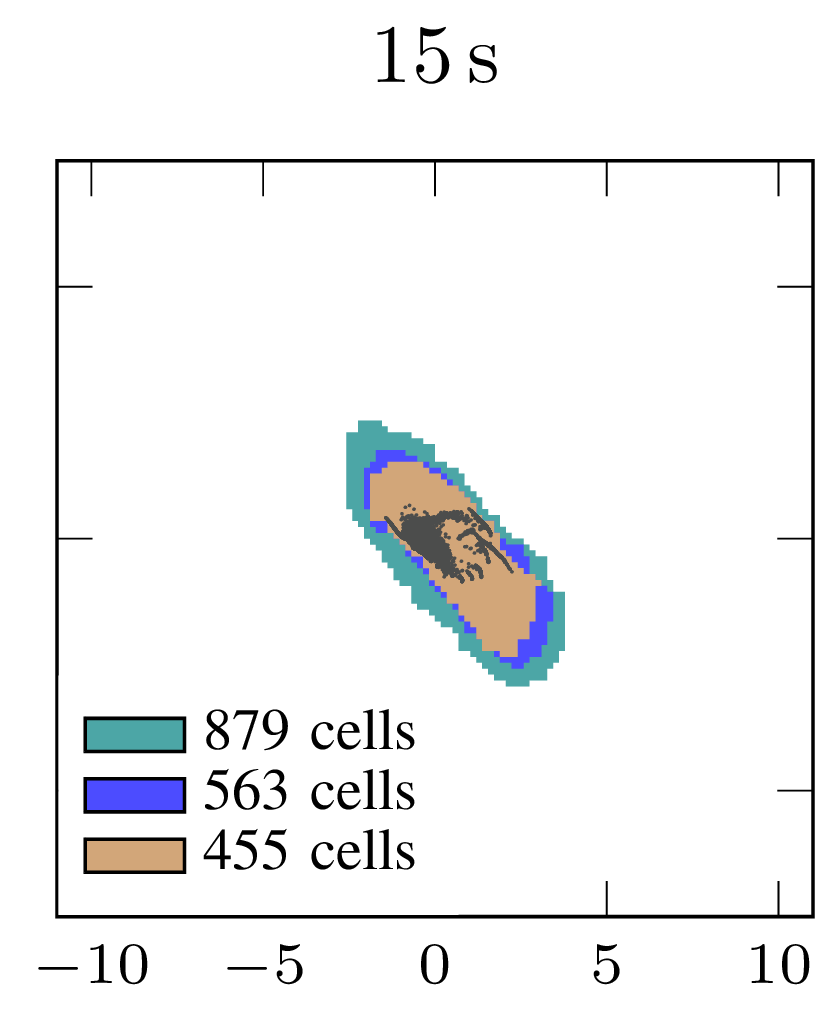}}
	\hspace{-1.1em}
	\subfloat{\includegraphics{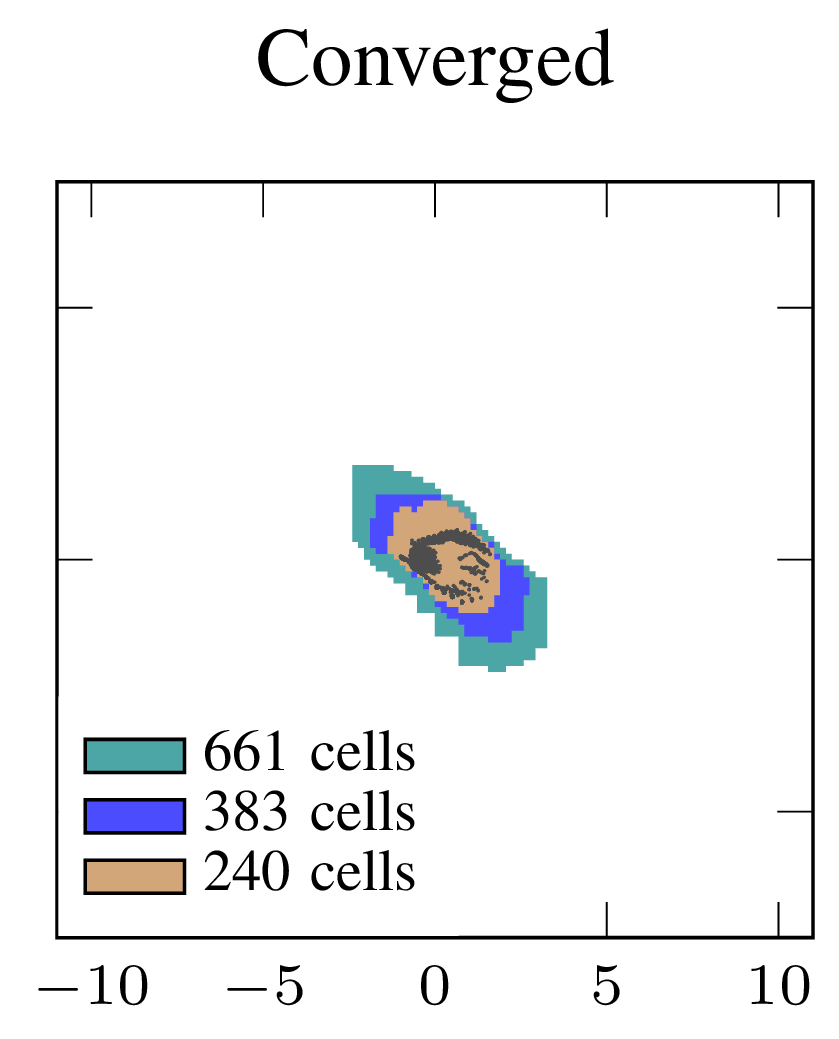}}
	\hspace{-1.5em}
	\caption{Reachable sets over time using three different methods when starting from $p\in[-\qty{10}{\meter}, \qty{10}{\meter}]$ and $\theta\in[-\qty{10}{\degree}, \qty{10}{\degree}]$ in taxiing system. 
		The colors \textcolor{black!70}{black}, \textcolor{teal!70}{teal}, 
		\textcolor{blue!70}{blue}, and \textcolor{brown!70}{brown}
		represent simulations, baseline, $1$-step, and $2$-step methods, respectively.}
	\label{fig:taxi_system_reachable_sets_original}
\end{figure*}
The multi-step unrolling strategy repeats the composition and propagates the resulting star sets across multiple steps.
As shown in Fig.~\ref{fig:taxi_system_single_cell_propogate}, 
the $2$-step method (row 3, column 3) at step $2$ significantly reduces overapproximation compared to the $1$-step method (row 2, column 3), primarily by mitigating multi-step error.
Consequently, comparing the baseline, $1$-step, and $2$-step methods at step $2$, the number of reachable cells decreases from $55$ to $40$ to $25$, with reachable sets becoming closer to simulation results.
%
As the number of steps $m$ increases, the verification task become more complex to the growing scale of network architectures and expanding input dimensions with two additional latent variables at each step.
In this experiment, we consider $m=1, 2$.


Two properties are evaluated for the closed-loop system:
\begin{itemize}[leftmargin=*]
	\item P1: The aircraft remains on the runway, ensuring that the magnitude of the crosstrack error does not exceed $\qty{10}{\meter}$.
	\item P2: The aircraft is steered toward the runway center, leading to convergence of the reachable set to an invariant set.
\end{itemize}
We begin by evaluating P2 using the same initial set as prior work,
where $p\in[\qty{-10}{\meter}, \qty{10}{\meter}]$ and $\theta\in[\qty{-10}{\degree}, \qty{10}{\degree}]$. 
The simulations and reachable sets computed using the baseline, $1$-step, and $2$-step methods are depicted in Fig.~\ref{fig:taxi_system_reachable_sets_original}.
Over time, the reachable sets of all three methods gradually contract and 
eventually converge to invariant sets at $\qty{22}{\second}$, $\qty{22}{\second}$, and $\qty{26}{\second}$, respectively. This demonstrates P2, showing that the aircraft, guided by the image-based controller, converges near the runway center.
Further, the reachable sets computed using the baseline, $1$-step, and $2$-step methods progressively decrease and align more closely with the simulations. 
Upon convergence, the number of reachable cells for three methods are $661$, $383$, and $240$, respectively.
The steady state area with the baseline method is $175$\% larger than our two-step approach. 
These numerical results show that the proposed methods significantly reduce the overapproximation compared 
to the baseline.
%
%
The importance of our accuracy improvements becomes more apparent when parameters of the case study are altered, detailed in the Appendix \ref{appendix:aats}.

%% file: experiments-advanced_emergency_braking_system.tex
\subsection{Advanced Emergency Braking System}
\label{ssec:carla}

We next evaluate the proposed approach using an advanced emergency braking system~\cite{cai2020iccps} with an image-based controller,
demonstrating that our improvements permit the verification of image-based NNCS that are significantly more complex than the prior work.
%


\subsubsection{System Details}
The system is to apply braking force to safely stop the host vehicle when approaching a stopped vehicle ahead. 
The state is defined by the distance to the obstacle $d$ 
and the host vehicle's velocity $v$, evolving according to linear dynamics: 
$d_{k+1} = d_k - v_k \Delta t$ and $v_{k+1} = v_k - a_k \Delta t$,
%
where time step $\Delta t = \qty{0.05}{\second}$.
The deceleration $a_k$ is calculated as $a_k = 0.009 u_k + 0.0042$, with $u_k$ representing the braking force predicted by an image-based controller.
%

We consider two versions of an image-based controller: 
one using a convolutional network and the other a transformer architecture. 
The perception system is abstracted using a cGAN, 
which incorporates the distance $d$ as a conditional variable, along with four latent variables, each bounded by $\pm 10^{-2}$. 
The cGAN also comes in two network variants, aligning with the controller versions.
%
The convolutional variant comprises a total of $8$ convolutional layers, including both cGAN and controller, while the transformer variant includes a total of $22$ convolutional layers and $2$ self-attention layers.
Detailed network architectures and the generated images are available in Appendix~\ref{appendix:aebs}.

\subsubsection{Verification Results}

\begin{figure}[!t]
	\centering
	\captionsetup[subfloat]{labelfont=footnotesize,textfont=scriptsize,position=top,margin={0.7cm,0.0cm}, skip=-3pt}
	\tikzsetnextfilename{aebs_backward_conv_20hz}%
	\subfloat[Convolutional ($\qty{20}{\hertz}$)]{\label{fig:conv_20}\input{figures/aebs_backward_conv_20hz}}\hspace{-0.2em}
	%
	%
	\captionsetup[subfloat]{labelfont=footnotesize,textfont=scriptsize,position=top, margin={0cm,0cm}}
	\tikzsetnextfilename{aebs_backward_conv_10hz}%
	\subfloat[Convolutional ($\qty{10}{\hertz}$)]{\label{fig:conv_10}\input{figures/aebs_backward_conv_10hz}}\hspace{-0.2em}
	
	\captionsetup[subfloat]{labelfont=footnotesize,textfont=scriptsize,position=top,margin={0.7cm,0.0cm}, skip=-3pt}
	\tikzsetnextfilename{aebs_backward_conv_5hz}%
	\subfloat[Convolutional ($\qty{5}{\hertz}$)]{\label{fig:conv_5}\input{figures/aebs_backward_conv_5hz}}\hspace{-0.2em}
	%
	%
	\captionsetup[subfloat]{labelfont=footnotesize,textfont=scriptsize,position=top, margin={0cm,0cm}}
	\tikzsetnextfilename{aebs_backward_vit_10hz}%
	\subfloat[Transformer ($\qty{10}{\hertz}$)]{\label{fig:vit_10}\input{figures/aebs_backward_vit_10hz}}
	\caption{
		The state sets guaranteed to be safe 
		for different controllers in braking system. 
		Cells are identified as unsafe by  simulations are colored in \textcolor[rgb]{0.98,0.50,0.44}{red},
		and cells are verified as safe using $1$-, $2$-, and $3$-step methods are in \textcolor[rgb]{0.56,0.81,0.81}{green}, \textcolor[rgb]{0.51,0.69,0.82}{blue}, and \textcolor[rgb]{0.75,0.72,0.86}{purple}.
		Quantitative results are presented in Appendix~\ref{appendix:aebs}.
		}
	\label{fig:aebs_backward}
\end{figure}

The state space with $d \in [\qty{0}{\meter}, \qty{60}{\meter}]$ and $v \in [\qty[per-mode=symbol]{0}{\meter\per\second}, \qty[per-mode=symbol]{30}{\meter\per\second}]$
is divided into a grid of $100\times100$ equal-size cells, 
while the latent space is not divided.
Given the extensive scale and the non-$\mathsf{ReLU}$ activation functions in the network, performing composition with the exact analysis approach is infeasible. 
Instead, we perform composition the other way, by appending the linear dynamics 
to the neural networks as an additional layer and then running range analysis with $\alpha$-$\beta$-CROWN.  
%
Since the state space is pre-divided, 
the process of abstracting reachable sets into cells can be integrated into the verification process. 
For each output, we begin with simulations to determine the minimized range of the output, then identify the lower and upper indices of cells where all simulations are located.
We incrementally expand the indices until the output is proved to be safe within the range.  
This approach allows us to effectively compute the reachable set represented by interval bounds.

For this system, we consider a single specification:
\begin{itemize}[leftmargin=*]
	\item P1: The host vehicle comes to a stop before colliding with the lead, meaning that the distance to the leading vehicle should never reach $\qty{0}{\meter}$ before 
the velocity reaches $\qty[per-mode=symbol]{0}{\meter\per\second}$.
\end{itemize}

\begin{figure}[!t]
	\centering
	\tikzsetnextfilename{taxi_system_compare_1_step_and_2_step}
	\includegraphics[width=1.0\columnwidth]{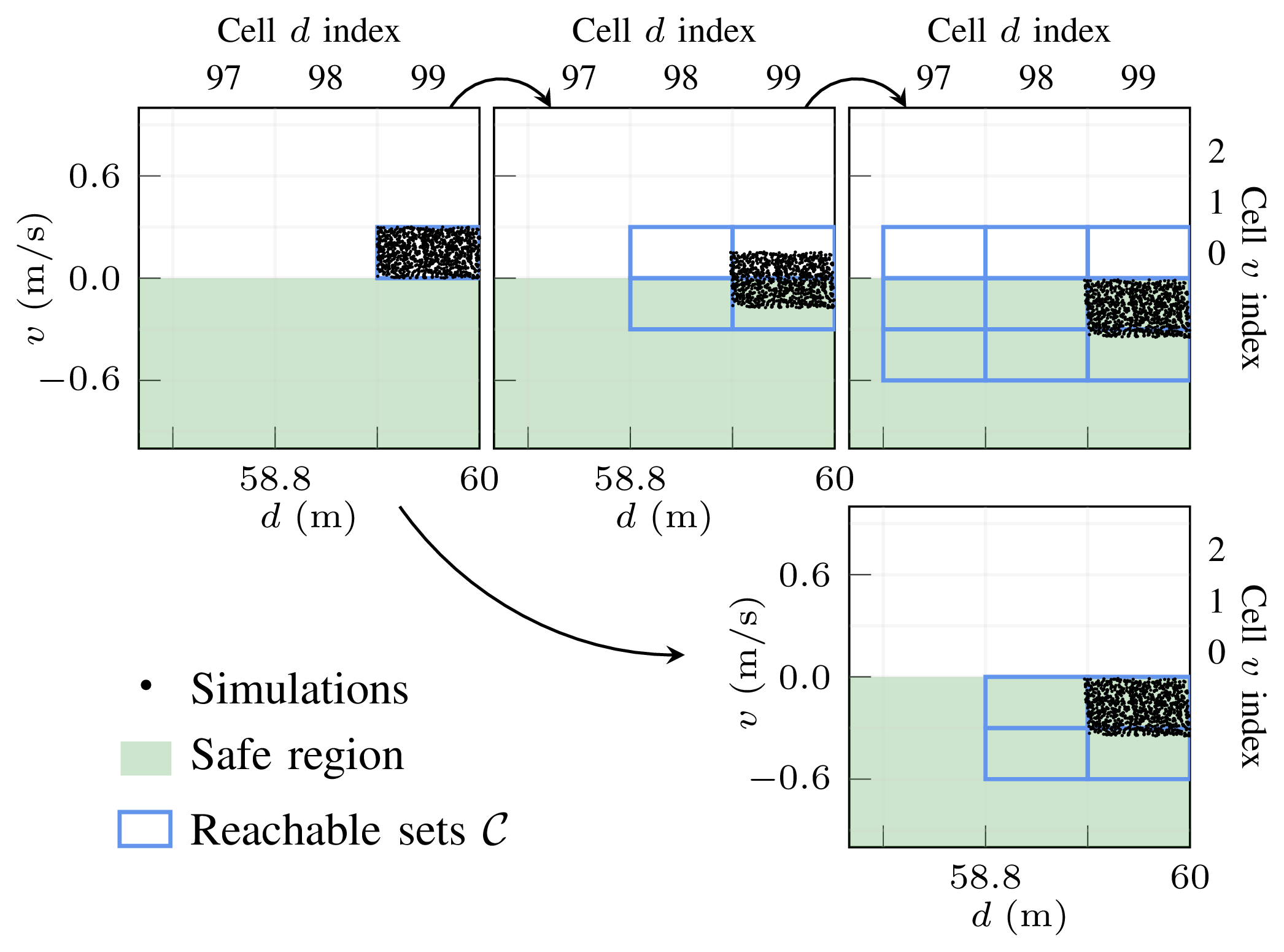}
	\caption{Comparison of the reachable sets from a single cell computed by $1$-step (first row) and $2$-step methods (second row) in the braking system. }
	\label{fig:taxi_system_compare_1_step_and_2_step}
\end{figure}

First, we analyze the image-based controller using the convolutional network variant.
We run $5000$ simulations starting from random points in each cell, in order to estimate the number of possible safe cells.
In Fig.~\ref{fig:conv_20}, 
$3463$ cells within the state space are detected as unsafe using simulations,
while the remaining cells are candidate safe cells.
We unroll the system with $m=1, 2$, and $3$ steps, and run backward reachability analysis to show which states can be proven safe in Fig.~\ref{fig:conv_20}.
%
Surprisingly, no states can be proven safe using the $1$-step method, even in situations with large distances and nearly-zero velocities. 
Compared with the aircraft taxiing system, the dynamics in this system moves the states significantly less at each control cycle, leading to excessive multi-step error from the abstraction process.
This is illustrated and detailed discussed in 
the first row of Fig.~\ref{fig:taxi_system_compare_1_step_and_2_step}.
Starting from cell $(99, 0)$, the $1$-step method extends the reachable set a cell located on the left, $(98, 0)$ after one step. This trend continues, eventually extending reachable set to the leftmost cell where an unsafe state is reached.
This experiment also demonstrates that the baseline method is ineffective in verifying such a system, given the $1$-step method has lower overapproximation compared to the baseline approach.
However, as shown in the second row, 
the $2$-step method can prove the safety of cell $(99, 0)$.

As the unrolling steps $m$ increases,
the number of provably safe cells also increases,
with $2$-step and $3$-step methods verifying $384$ and $2669$ cells, respectively.
However, 
a substantial number of $3868$ cells remain inconclusive. 
This occurs due to the control cycle's period of $\qty{0.05}{\second}$, where even $3$-step analysis has substantial multi-step error.
Unrolling beyond $3$ steps becomes practically difficult as the composed neural network increases in size, leading to an increase analysis time with the $\alpha$-$\beta$-CROWN verification tool.
Detailed results of runtime is provided in the Appendix \ref{appendix:run_time}.
%
%
%
%

As the primary cause of inconclusive cells is multi-step error from the abstraction process, we
can consider decreasing the control frequency to reduce this effect.
This is counterintuitive---a lower frequency should lead to \emph{worse} control performance and therefore more unsafe cells. However, due to reduced multi-step error this could actually allow the method to verify more cells as provably safe.
We analyze the system by reducing the control 
frequency from $\qty{20}{\hertz}$ to $\qty{10}{\hertz}$ (Fig.~\ref{fig:conv_10}) and
$\qty{5}{\hertz}$ (Fig.~\ref{fig:conv_5}).
As expected, the control performance becomes slightly worse as the control frequency decreases, resulting in a minor increase in the number of unsafe cells identified through simulation, by $1.7\%$ for $\qty{10}{\hertz}$ and $4.4\%$ for $\qty{5}{\hertz}$.
However, the decrease in frequency improves the ability to establish provably safe cells for all methods compared to their original results---the white gap in the figure is reduced from $38.7\%$ of the cells with the original $\qty{20}{\hertz}$ frequency.
For the $\qty{10}{\hertz}$, with $3$-step analysis only $7.4\%$ of the cells remain inconclusive.
For the $\qty{5}{\hertz}$, this is further reduced to $3.5\%$ of the cells.



We also evaluate the controller with transformer layers (Fig.~\ref{fig:vit_10}), where simulations identify $3517$ unsafe cells. As $m$ increases,
the numbers of verified safe cells increase, reaching $390$, $2890$, and $5707$ using  $1$-, $2$-, and $3$-step methods, respectively, while $7.8\%$ of cells  remain inconclusive.





%% file: figures/aebs_backward_conv_20hz.tex
\begin{tikzpicture}

	\definecolor{lightgray204}{RGB}{204,204,204}
	\definecolor{red}{RGB}{250,127,111}
	\definecolor{green}{RGB}{142,207,201}
	\definecolor{yellow}{RGB}{190,184,220}
	\definecolor{blue}{RGB}{130,176,210}
	
	\pgfplotsset{compat = newest,		
		scale only axis,
		width=3.0cm,
		height=3.0cm,
		every tick label/.append style={font=\scriptsize},
	}
		\begin{axis}[
		tick pos=left,
		xmin=0, xmax=60,
		ymin=0, ymax=30,
		xlabel style={font=\scriptsize, yshift=5pt},
		ylabel={$v\ (\unit{\meter\per\second})$},
		ylabel style={font=\scriptsize},
		title style={font=\scriptsize},
		]
		\path [draw=none, fill=red]
		(axis cs:52.2,30.001)
		--(axis cs:52.201,30.001)
		--(axis cs:52.8,30.001)
		--(axis cs:52.801,30.001)
		--(axis cs:53.4,30.001)
		--(axis cs:53.401,30.001)
		--(axis cs:54,30.001)
		--(axis cs:54.001,30.001)
		--(axis cs:54.6,30.001)
		--(axis cs:54.601,30.001)
		--(axis cs:55.2,30.001)
		--(axis cs:55.201,30.001)
		--(axis cs:55.8,30.001)
		--(axis cs:55.801,30.001)
		--(axis cs:56.4,30.001)
		--(axis cs:56.401,30.001)
		--(axis cs:57,30.001)
		--(axis cs:57.001,30.001)
		--(axis cs:57.6,30.001)
		--(axis cs:57.601,30.001)
		--(axis cs:58.2,30.001)
		--(axis cs:58.201,30.001)
		--(axis cs:58.801,30.001)
		--(axis cs:58.801,29.7)
		--(axis cs:58.201,29.7)
		--(axis cs:58.2,29.7)
		--(axis cs:57.601,29.7)
		--(axis cs:57.601,29.4)
		--(axis cs:57.001,29.4)
		--(axis cs:57,29.4)
		--(axis cs:56.401,29.4)
		--(axis cs:56.401,29.1)
		--(axis cs:55.801,29.1)
		--(axis cs:55.8,29.1)
		--(axis cs:55.201,29.1)
		--(axis cs:55.201,28.8)
		--(axis cs:54.601,28.8)
		--(axis cs:54.601,28.5)
		--(axis cs:54.001,28.5)
		--(axis cs:54,28.5)
		--(axis cs:53.401,28.5)
		--(axis cs:53.401,28.2)
		--(axis cs:52.801,28.2)
		--(axis cs:52.8,28.2)
		--(axis cs:52.201,28.2)
		--(axis cs:52.201,27.9)
		--(axis cs:51.601,27.9)
		--(axis cs:51.6,27.9)
		--(axis cs:51.001,27.9)
		--(axis cs:51.001,27.6)
		--(axis cs:50.401,27.6)
		--(axis cs:50.4,27.6)
		--(axis cs:49.801,27.6)
		--(axis cs:49.801,27.3)
		--(axis cs:49.201,27.3)
		--(axis cs:49.201,27)
		--(axis cs:48.601,27)
		--(axis cs:48.6,27)
		--(axis cs:48.001,27)
		--(axis cs:48.001,26.7)
		--(axis cs:47.401,26.7)
		--(axis cs:47.4,26.7)
		--(axis cs:46.801,26.7)
		--(axis cs:46.801,26.4)
		--(axis cs:46.201,26.4)
		--(axis cs:46.201,26.1)
		--(axis cs:45.601,26.1)
		--(axis cs:45.6,26.1)
		--(axis cs:45.001,26.1)
		--(axis cs:45.001,25.8)
		--(axis cs:44.401,25.8)
		--(axis cs:44.4,25.8)
		--(axis cs:43.801,25.8)
		--(axis cs:43.801,25.5)
		--(axis cs:43.201,25.5)
		--(axis cs:43.201,25.2)
		--(axis cs:42.601,25.2)
		--(axis cs:42.6,25.2)
		--(axis cs:42.001,25.2)
		--(axis cs:42.001,24.9)
		--(axis cs:41.401,24.9)
		--(axis cs:41.4,24.9)
		--(axis cs:40.801,24.9)
		--(axis cs:40.801,24.6)
		--(axis cs:40.201,24.6)
		--(axis cs:40.201,24.3)
		--(axis cs:39.601,24.3)
		--(axis cs:39.6,24.3)
		--(axis cs:39.001,24.3)
		--(axis cs:39.001,24)
		--(axis cs:38.401,24)
		--(axis cs:38.401,23.7)
		--(axis cs:37.801,23.7)
		--(axis cs:37.8,23.7)
		--(axis cs:37.201,23.7)
		--(axis cs:37.201,23.4)
		--(axis cs:36.601,23.4)
		--(axis cs:36.601,23.1)
		--(axis cs:36.001,23.1)
		--(axis cs:36,23.1)
		--(axis cs:35.401,23.1)
		--(axis cs:35.401,22.8)
		--(axis cs:34.801,22.8)
		--(axis cs:34.801,22.5)
		--(axis cs:34.201,22.5)
		--(axis cs:34.2,22.5)
		--(axis cs:33.601,22.5)
		--(axis cs:33.601,22.2)
		--(axis cs:33.001,22.2)
		--(axis cs:33.001,21.9)
		--(axis cs:32.401,21.9)
		--(axis cs:32.401,21.6)
		--(axis cs:31.801,21.6)
		--(axis cs:31.8,21.6)
		--(axis cs:31.201,21.6)
		--(axis cs:31.201,21.3)
		--(axis cs:30.601,21.3)
		--(axis cs:30.601,21)
		--(axis cs:30.001,21)
		--(axis cs:30,21)
		--(axis cs:29.401,21)
		--(axis cs:29.401,20.7)
		--(axis cs:28.801,20.7)
		--(axis cs:28.801,20.4)
		--(axis cs:28.201,20.4)
		--(axis cs:28.201,20.1)
		--(axis cs:27.601,20.1)
		--(axis cs:27.6,20.1)
		--(axis cs:27.001,20.1)
		--(axis cs:27.001,19.8)
		--(axis cs:26.401,19.8)
		--(axis cs:26.401,19.5)
		--(axis cs:25.801,19.5)
		--(axis cs:25.801,19.2)
		--(axis cs:25.201,19.2)
		--(axis cs:25.201,18.9)
		--(axis cs:24.601,18.9)
		--(axis cs:24.6,18.9)
		--(axis cs:24.001,18.9)
		--(axis cs:24.001,18.6)
		--(axis cs:23.401,18.6)
		--(axis cs:23.401,18.3)
		--(axis cs:22.801,18.3)
		--(axis cs:22.801,18)
		--(axis cs:22.201,18)
		--(axis cs:22.201,17.7)
		--(axis cs:21.601,17.7)
		--(axis cs:21.601,17.4)
		--(axis cs:21.001,17.4)
		--(axis cs:21.001,17.1)
		--(axis cs:20.401,17.1)
		--(axis cs:20.401,16.8)
		--(axis cs:19.801,16.8)
		--(axis cs:19.801,16.5)
		--(axis cs:19.201,16.5)
		--(axis cs:19.2,16.5)
		--(axis cs:18.601,16.5)
		--(axis cs:18.601,16.2)
		--(axis cs:18.001,16.2)
		--(axis cs:18.001,15.9)
		--(axis cs:17.401,15.9)
		--(axis cs:17.401,15.6)
		--(axis cs:16.801,15.6)
		--(axis cs:16.801,15.3)
		--(axis cs:16.201,15.3)
		--(axis cs:16.201,15)
		--(axis cs:15.601,15)
		--(axis cs:15.601,14.7)
		--(axis cs:15.001,14.7)
		--(axis cs:15.001,14.401)
		--(axis cs:15.001,14.4)
		--(axis cs:15.001,14.1)
		--(axis cs:14.401,14.1)
		--(axis cs:14.401,13.8)
		--(axis cs:13.801,13.8)
		--(axis cs:13.801,13.5)
		--(axis cs:13.201,13.5)
		--(axis cs:13.201,13.2)
		--(axis cs:12.601,13.2)
		--(axis cs:12.601,12.9)
		--(axis cs:12.001,12.9)
		--(axis cs:12.001,12.6)
		--(axis cs:11.401,12.6)
		--(axis cs:11.401,12.301)
		--(axis cs:11.401,12.3)
		--(axis cs:11.401,12)
		--(axis cs:10.801,12)
		--(axis cs:10.801,11.7)
		--(axis cs:10.201,11.7)
		--(axis cs:10.201,11.4)
		--(axis cs:9.601,11.4)
		--(axis cs:9.601,11.101)
		--(axis cs:9.601,11.1)
		--(axis cs:9.601,10.8)
		--(axis cs:9.001,10.8)
		--(axis cs:9.001,10.5)
		--(axis cs:8.401,10.5)
		--(axis cs:8.401,10.201)
		--(axis cs:8.401,10.2)
		--(axis cs:8.401,9.9)
		--(axis cs:7.801,9.9)
		--(axis cs:7.801,9.6)
		--(axis cs:7.201,9.6)
		--(axis cs:7.201,9.301)
		--(axis cs:7.201,9.3)
		--(axis cs:7.201,9)
		--(axis cs:6.601,9)
		--(axis cs:6.601,8.7)
		--(axis cs:6.001,8.7)
		--(axis cs:6.001,8.401)
		--(axis cs:6.001,8.4)
		--(axis cs:6.001,8.1)
		--(axis cs:5.401,8.1)
		--(axis cs:5.401,7.801)
		--(axis cs:5.401,7.8)
		--(axis cs:5.401,7.5)
		--(axis cs:4.801,7.5)
		--(axis cs:4.801,7.201)
		--(axis cs:4.801,7.2)
		--(axis cs:4.801,6.9)
		--(axis cs:4.201,6.9)
		--(axis cs:4.201,6.601)
		--(axis cs:4.201,6.6)
		--(axis cs:4.201,6.301)
		--(axis cs:4.201,6.3)
		--(axis cs:4.201,6)
		--(axis cs:3.601,6)
		--(axis cs:3.601,5.701)
		--(axis cs:3.601,5.7)
		--(axis cs:3.601,5.4)
		--(axis cs:3.001,5.4)
		--(axis cs:3.001,5.101)
		--(axis cs:3.001,5.1)
		--(axis cs:3.001,4.801)
		--(axis cs:3.001,4.8)
		--(axis cs:3.001,4.5)
		--(axis cs:2.401,4.5)
		--(axis cs:2.401,4.201)
		--(axis cs:2.401,4.2)
		--(axis cs:2.401,3.9)
		--(axis cs:1.801,3.9)
		--(axis cs:1.801,3.601)
		--(axis cs:1.801,3.6)
		--(axis cs:1.801,3.301)
		--(axis cs:1.801,3.3)
		--(axis cs:1.801,3)
		--(axis cs:1.201,3)
		--(axis cs:1.201,2.701)
		--(axis cs:1.201,2.7)
		--(axis cs:1.201,2.401)
		--(axis cs:1.201,2.4)
		--(axis cs:1.201,2.1)
		--(axis cs:0.601,2.1)
		--(axis cs:0.601,1.801)
		--(axis cs:0.601,1.8)
		--(axis cs:0.601,1.501)
		--(axis cs:0.601,1.5)
		--(axis cs:0.601,1.201)
		--(axis cs:0.601,1.2)
		--(axis cs:0.601,0.901)
		--(axis cs:0.601,0.9)
		--(axis cs:0.601,0.601)
		--(axis cs:0.601,0.6)
		--(axis cs:0.601,0.301)
		--(axis cs:0.601,0.3)
		--(axis cs:0.601,0)
		--(axis cs:0,0)
		--(axis cs:0,0.3)
		--(axis cs:0,0.301)
		--(axis cs:0,0.6)
		--(axis cs:0,0.601)
		--(axis cs:0,0.9)
		--(axis cs:0,0.901)
		--(axis cs:0,1.2)
		--(axis cs:0,1.201)
		--(axis cs:0,1.5)
		--(axis cs:0,1.501)
		--(axis cs:0,1.8)
		--(axis cs:0,1.801)
		--(axis cs:0,2.1)
		--(axis cs:0,2.101)
		--(axis cs:0,2.4)
		--(axis cs:0,2.401)
		--(axis cs:0,2.7)
		--(axis cs:0,2.701)
		--(axis cs:0,3)
		--(axis cs:0,3.001)
		--(axis cs:0,3.3)
		--(axis cs:0,3.301)
		--(axis cs:0,3.6)
		--(axis cs:0,3.601)
		--(axis cs:0,3.9)
		--(axis cs:0,3.901)
		--(axis cs:0,4.2)
		--(axis cs:0,4.201)
		--(axis cs:0,4.5)
		--(axis cs:0,4.501)
		--(axis cs:0,4.8)
		--(axis cs:0,4.801)
		--(axis cs:0,5.1)
		--(axis cs:0,5.101)
		--(axis cs:0,5.4)
		--(axis cs:0,5.401)
		--(axis cs:0,5.7)
		--(axis cs:0,5.701)
		--(axis cs:0,6)
		--(axis cs:0,6.001)
		--(axis cs:0,6.3)
		--(axis cs:0,6.301)
		--(axis cs:0,6.6)
		--(axis cs:0,6.601)
		--(axis cs:0,6.9)
		--(axis cs:0,6.901)
		--(axis cs:0,7.2)
		--(axis cs:0,7.201)
		--(axis cs:0,7.5)
		--(axis cs:0,7.501)
		--(axis cs:0,7.8)
		--(axis cs:0,7.801)
		--(axis cs:0,8.1)
		--(axis cs:0,8.101)
		--(axis cs:0,8.4)
		--(axis cs:0,8.401)
		--(axis cs:0,8.7)
		--(axis cs:0,8.701)
		--(axis cs:0,9)
		--(axis cs:0,9.001)
		--(axis cs:0,9.3)
		--(axis cs:0,9.301)
		--(axis cs:0,9.6)
		--(axis cs:0,9.601)
		--(axis cs:0,9.9)
		--(axis cs:0,9.901)
		--(axis cs:0,10.2)
		--(axis cs:0,10.201)
		--(axis cs:0,10.5)
		--(axis cs:0,10.501)
		--(axis cs:0,10.8)
		--(axis cs:0,10.801)
		--(axis cs:0,11.1)
		--(axis cs:0,11.101)
		--(axis cs:0,11.4)
		--(axis cs:0,11.401)
		--(axis cs:0,11.7)
		--(axis cs:0,11.701)
		--(axis cs:0,12)
		--(axis cs:0,12.001)
		--(axis cs:0,12.3)
		--(axis cs:0,12.301)
		--(axis cs:0,12.6)
		--(axis cs:0,12.601)
		--(axis cs:0,12.9)
		--(axis cs:0,12.901)
		--(axis cs:0,13.2)
		--(axis cs:0,13.201)
		--(axis cs:0,13.5)
		--(axis cs:0,13.501)
		--(axis cs:0,13.8)
		--(axis cs:0,13.801)
		--(axis cs:0,14.1)
		--(axis cs:0,14.101)
		--(axis cs:0,14.4)
		--(axis cs:0,14.401)
		--(axis cs:0,14.7)
		--(axis cs:0,14.701)
		--(axis cs:0,15)
		--(axis cs:0,15.001)
		--(axis cs:0,15.3)
		--(axis cs:0,15.301)
		--(axis cs:0,15.6)
		--(axis cs:0,15.601)
		--(axis cs:0,15.9)
		--(axis cs:0,15.901)
		--(axis cs:0,16.2)
		--(axis cs:0,16.201)
		--(axis cs:0,16.5)
		--(axis cs:0,16.501)
		--(axis cs:0,16.8)
		--(axis cs:0,16.801)
		--(axis cs:0,17.1)
		--(axis cs:0,17.101)
		--(axis cs:0,17.4)
		--(axis cs:0,17.401)
		--(axis cs:0,17.7)
		--(axis cs:0,17.701)
		--(axis cs:0,18)
		--(axis cs:0,18.001)
		--(axis cs:0,18.3)
		--(axis cs:0,18.301)
		--(axis cs:0,18.6)
		--(axis cs:0,18.601)
		--(axis cs:0,18.9)
		--(axis cs:0,18.901)
		--(axis cs:0,19.2)
		--(axis cs:0,19.201)
		--(axis cs:0,19.5)
		--(axis cs:0,19.501)
		--(axis cs:0,19.8)
		--(axis cs:0,19.801)
		--(axis cs:0,20.1)
		--(axis cs:0,20.101)
		--(axis cs:0,20.4)
		--(axis cs:0,20.401)
		--(axis cs:0,20.7)
		--(axis cs:0,20.701)
		--(axis cs:0,21)
		--(axis cs:0,21.001)
		--(axis cs:0,21.3)
		--(axis cs:0,21.301)
		--(axis cs:0,21.6)
		--(axis cs:0,21.601)
		--(axis cs:0,21.9)
		--(axis cs:0,21.901)
		--(axis cs:0,22.2)
		--(axis cs:0,22.201)
		--(axis cs:0,22.5)
		--(axis cs:0,22.501)
		--(axis cs:0,22.8)
		--(axis cs:0,22.801)
		--(axis cs:0,23.1)
		--(axis cs:0,23.101)
		--(axis cs:0,23.4)
		--(axis cs:0,23.401)
		--(axis cs:0,23.7)
		--(axis cs:0,23.701)
		--(axis cs:0,24)
		--(axis cs:0,24.001)
		--(axis cs:0,24.3)
		--(axis cs:0,24.301)
		--(axis cs:0,24.6)
		--(axis cs:0,24.601)
		--(axis cs:0,24.9)
		--(axis cs:0,24.901)
		--(axis cs:0,25.2)
		--(axis cs:0,25.201)
		--(axis cs:0,25.5)
		--(axis cs:0,25.501)
		--(axis cs:0,25.8)
		--(axis cs:0,25.801)
		--(axis cs:0,26.1)
		--(axis cs:0,26.101)
		--(axis cs:0,26.4)
		--(axis cs:0,26.401)
		--(axis cs:0,26.7)
		--(axis cs:0,26.701)
		--(axis cs:0,27)
		--(axis cs:0,27.001)
		--(axis cs:0,27.3)
		--(axis cs:0,27.301)
		--(axis cs:0,27.6)
		--(axis cs:0,27.601)
		--(axis cs:0,27.9)
		--(axis cs:0,27.901)
		--(axis cs:0,28.2)
		--(axis cs:0,28.201)
		--(axis cs:0,28.5)
		--(axis cs:0,28.501)
		--(axis cs:0,28.8)
		--(axis cs:0,28.801)
		--(axis cs:0,29.1)
		--(axis cs:0,29.101)
		--(axis cs:0,29.4)
		--(axis cs:0,29.401)
		--(axis cs:0,29.7)
		--(axis cs:0,29.701)
		--(axis cs:0,30.001)
		--(axis cs:0.6,30.001)
		--(axis cs:0.601,30.001)
		--(axis cs:1.2,30.001)
		--(axis cs:1.201,30.001)
		--(axis cs:1.8,30.001)
		--(axis cs:1.801,30.001)
		--(axis cs:2.4,30.001)
		--(axis cs:2.401,30.001)
		--(axis cs:3,30.001)
		--(axis cs:3.001,30.001)
		--(axis cs:3.6,30.001)
		--(axis cs:3.601,30.001)
		--(axis cs:4.2,30.001)
		--(axis cs:4.201,30.001)
		--(axis cs:4.8,30.001)
		--(axis cs:4.801,30.001)
		--(axis cs:5.4,30.001)
		--(axis cs:5.401,30.001)
		--(axis cs:6,30.001)
		--(axis cs:6.001,30.001)
		--(axis cs:6.6,30.001)
		--(axis cs:6.601,30.001)
		--(axis cs:7.2,30.001)
		--(axis cs:7.201,30.001)
		--(axis cs:7.8,30.001)
		--(axis cs:7.801,30.001)
		--(axis cs:8.4,30.001)
		--(axis cs:8.401,30.001)
		--(axis cs:9,30.001)
		--(axis cs:9.001,30.001)
		--(axis cs:9.6,30.001)
		--(axis cs:9.601,30.001)
		--(axis cs:10.2,30.001)
		--(axis cs:10.201,30.001)
		--(axis cs:10.8,30.001)
		--(axis cs:10.801,30.001)
		--(axis cs:11.4,30.001)
		--(axis cs:11.401,30.001)
		--(axis cs:12,30.001)
		--(axis cs:12.001,30.001)
		--(axis cs:12.6,30.001)
		--(axis cs:12.601,30.001)
		--(axis cs:13.2,30.001)
		--(axis cs:13.201,30.001)
		--(axis cs:13.8,30.001)
		--(axis cs:13.801,30.001)
		--(axis cs:14.4,30.001)
		--(axis cs:14.401,30.001)
		--(axis cs:15,30.001)
		--(axis cs:15.001,30.001)
		--(axis cs:15.6,30.001)
		--(axis cs:15.601,30.001)
		--(axis cs:16.2,30.001)
		--(axis cs:16.201,30.001)
		--(axis cs:16.8,30.001)
		--(axis cs:16.801,30.001)
		--(axis cs:17.4,30.001)
		--(axis cs:17.401,30.001)
		--(axis cs:18,30.001)
		--(axis cs:18.001,30.001)
		--(axis cs:18.6,30.001)
		--(axis cs:18.601,30.001)
		--(axis cs:19.2,30.001)
		--(axis cs:19.201,30.001)
		--(axis cs:19.8,30.001)
		--(axis cs:19.801,30.001)
		--(axis cs:20.4,30.001)
		--(axis cs:20.401,30.001)
		--(axis cs:21,30.001)
		--(axis cs:21.001,30.001)
		--(axis cs:21.6,30.001)
		--(axis cs:21.601,30.001)
		--(axis cs:22.2,30.001)
		--(axis cs:22.201,30.001)
		--(axis cs:22.8,30.001)
		--(axis cs:22.801,30.001)
		--(axis cs:23.4,30.001)
		--(axis cs:23.401,30.001)
		--(axis cs:24,30.001)
		--(axis cs:24.001,30.001)
		--(axis cs:24.6,30.001)
		--(axis cs:24.601,30.001)
		--(axis cs:25.2,30.001)
		--(axis cs:25.201,30.001)
		--(axis cs:25.8,30.001)
		--(axis cs:25.801,30.001)
		--(axis cs:26.4,30.001)
		--(axis cs:26.401,30.001)
		--(axis cs:27,30.001)
		--(axis cs:27.001,30.001)
		--(axis cs:27.6,30.001)
		--(axis cs:27.601,30.001)
		--(axis cs:28.2,30.001)
		--(axis cs:28.201,30.001)
		--(axis cs:28.8,30.001)
		--(axis cs:28.801,30.001)
		--(axis cs:29.4,30.001)
		--(axis cs:29.401,30.001)
		--(axis cs:30,30.001)
		--(axis cs:30.001,30.001)
		--(axis cs:30.6,30.001)
		--(axis cs:30.601,30.001)
		--(axis cs:31.2,30.001)
		--(axis cs:31.201,30.001)
		--(axis cs:31.8,30.001)
		--(axis cs:31.801,30.001)
		--(axis cs:32.4,30.001)
		--(axis cs:32.401,30.001)
		--(axis cs:33,30.001)
		--(axis cs:33.001,30.001)
		--(axis cs:33.6,30.001)
		--(axis cs:33.601,30.001)
		--(axis cs:34.2,30.001)
		--(axis cs:34.201,30.001)
		--(axis cs:34.8,30.001)
		--(axis cs:34.801,30.001)
		--(axis cs:35.4,30.001)
		--(axis cs:35.401,30.001)
		--(axis cs:36,30.001)
		--(axis cs:36.001,30.001)
		--(axis cs:36.6,30.001)
		--(axis cs:36.601,30.001)
		--(axis cs:37.2,30.001)
		--(axis cs:37.201,30.001)
		--(axis cs:37.8,30.001)
		--(axis cs:37.801,30.001)
		--(axis cs:38.4,30.001)
		--(axis cs:38.401,30.001)
		--(axis cs:39,30.001)
		--(axis cs:39.001,30.001)
		--(axis cs:39.6,30.001)
		--(axis cs:39.601,30.001)
		--(axis cs:40.2,30.001)
		--(axis cs:40.201,30.001)
		--(axis cs:40.8,30.001)
		--(axis cs:40.801,30.001)
		--(axis cs:41.4,30.001)
		--(axis cs:41.401,30.001)
		--(axis cs:42,30.001)
		--(axis cs:42.001,30.001)
		--(axis cs:42.6,30.001)
		--(axis cs:42.601,30.001)
		--(axis cs:43.2,30.001)
		--(axis cs:43.201,30.001)
		--(axis cs:43.8,30.001)
		--(axis cs:43.801,30.001)
		--(axis cs:44.4,30.001)
		--(axis cs:44.401,30.001)
		--(axis cs:45,30.001)
		--(axis cs:45.001,30.001)
		--(axis cs:45.6,30.001)
		--(axis cs:45.601,30.001)
		--(axis cs:46.2,30.001)
		--(axis cs:46.201,30.001)
		--(axis cs:46.8,30.001)
		--(axis cs:46.801,30.001)
		--(axis cs:47.4,30.001)
		--(axis cs:47.401,30.001)
		--(axis cs:48,30.001)
		--(axis cs:48.001,30.001)
		--(axis cs:48.6,30.001)
		--(axis cs:48.601,30.001)
		--(axis cs:49.2,30.001)
		--(axis cs:49.201,30.001)
		--(axis cs:49.8,30.001)
		--(axis cs:49.801,30.001)
		--(axis cs:50.4,30.001)
		--(axis cs:50.401,30.001)
		--(axis cs:51,30.001)
		--(axis cs:51.001,30.001)
		--(axis cs:51.6,30.001)
		--(axis cs:51.601,30.001)
		--cycle;
		\path [draw=none, fill=yellow]
		(axis cs:5.401,0)
		--(axis cs:5.4,0)
		--(axis cs:4.801,0)
		--(axis cs:4.8,0)
		--(axis cs:4.201,0)
		--(axis cs:4.2,0)
		--(axis cs:3.601,0)
		--(axis cs:3.6,0)
		--(axis cs:3.001,0)
		--(axis cs:3,0)
		--(axis cs:2.401,0)
		--(axis cs:2.4,0)
		--(axis cs:1.801,0)
		--(axis cs:1.8,0)
		--(axis cs:1.201,0)
		--(axis cs:1.2,0)
		--(axis cs:0.6,0)
		--(axis cs:0.6,0.301)
		--(axis cs:1.2,0.301)
		--(axis cs:1.201,0.301)
		--(axis cs:1.8,0.301)
		--(axis cs:1.8,0.601)
		--(axis cs:2.4,0.601)
		--(axis cs:2.4,0.901)
		--(axis cs:3,0.901)
		--(axis cs:3,1.201)
		--(axis cs:3.6,1.201)
		--(axis cs:3.6,1.501)
		--(axis cs:4.2,1.501)
		--(axis cs:4.201,1.501)
		--(axis cs:4.8,1.501)
		--(axis cs:4.8,1.801)
		--(axis cs:5.4,1.801)
		--(axis cs:5.4,2.101)
		--(axis cs:6,2.101)
		--(axis cs:6,2.401)
		--(axis cs:6.6,2.401)
		--(axis cs:6.6,2.701)
		--(axis cs:7.2,2.701)
		--(axis cs:7.2,3.001)
		--(axis cs:7.8,3.001)
		--(axis cs:7.8,3.301)
		--(axis cs:8.4,3.301)
		--(axis cs:8.4,3.601)
		--(axis cs:9,3.601)
		--(axis cs:9,3.901)
		--(axis cs:9.6,3.901)
		--(axis cs:9.601,3.901)
		--(axis cs:10.2,3.901)
		--(axis cs:10.2,4.201)
		--(axis cs:10.8,4.201)
		--(axis cs:10.801,4.201)
		--(axis cs:11.4,4.201)
		--(axis cs:11.4,4.501)
		--(axis cs:12,4.501)
		--(axis cs:12.001,4.501)
		--(axis cs:12.6,4.501)
		--(axis cs:12.6,4.801)
		--(axis cs:13.2,4.801)
		--(axis cs:13.201,4.801)
		--(axis cs:13.8,4.801)
		--(axis cs:13.8,5.101)
		--(axis cs:14.4,5.101)
		--(axis cs:14.401,5.101)
		--(axis cs:15,5.101)
		--(axis cs:15,5.401)
		--(axis cs:15.6,5.401)
		--(axis cs:15.601,5.401)
		--(axis cs:16.2,5.401)
		--(axis cs:16.2,5.701)
		--(axis cs:16.8,5.701)
		--(axis cs:16.801,5.701)
		--(axis cs:17.4,5.701)
		--(axis cs:17.4,6.001)
		--(axis cs:18,6.001)
		--(axis cs:18.001,6.001)
		--(axis cs:18.6,6.001)
		--(axis cs:18.6,6.301)
		--(axis cs:19.2,6.301)
		--(axis cs:19.201,6.301)
		--(axis cs:19.8,6.301)
		--(axis cs:19.8,6.601)
		--(axis cs:20.4,6.601)
		--(axis cs:20.401,6.601)
		--(axis cs:21,6.601)
		--(axis cs:21,6.901)
		--(axis cs:21.6,6.901)
		--(axis cs:21.601,6.901)
		--(axis cs:22.2,6.901)
		--(axis cs:22.2,7.201)
		--(axis cs:22.8,7.201)
		--(axis cs:22.801,7.201)
		--(axis cs:23.4,7.201)
		--(axis cs:23.4,7.501)
		--(axis cs:24,7.501)
		--(axis cs:24.001,7.501)
		--(axis cs:24.6,7.501)
		--(axis cs:24.6,7.801)
		--(axis cs:25.2,7.801)
		--(axis cs:25.201,7.801)
		--(axis cs:25.8,7.801)
		--(axis cs:25.8,8.101)
		--(axis cs:26.4,8.101)
		--(axis cs:26.401,8.101)
		--(axis cs:27,8.101)
		--(axis cs:27.001,8.101)
		--(axis cs:27.6,8.101)
		--(axis cs:27.6,8.401)
		--(axis cs:28.2,8.401)
		--(axis cs:28.201,8.401)
		--(axis cs:28.8,8.401)
		--(axis cs:28.801,8.401)
		--(axis cs:29.4,8.401)
		--(axis cs:29.4,8.701)
		--(axis cs:30,8.701)
		--(axis cs:30.001,8.701)
		--(axis cs:30.6,8.701)
		--(axis cs:30.601,8.701)
		--(axis cs:31.2,8.701)
		--(axis cs:31.2,9.001)
		--(axis cs:31.8,9.001)
		--(axis cs:31.801,9.001)
		--(axis cs:32.4,9.001)
		--(axis cs:32.401,9.001)
		--(axis cs:33,9.001)
		--(axis cs:33,9.301)
		--(axis cs:33.6,9.301)
		--(axis cs:33.601,9.301)
		--(axis cs:34.2,9.301)
		--(axis cs:34.201,9.301)
		--(axis cs:34.8,9.301)
		--(axis cs:34.8,9.601)
		--(axis cs:35.4,9.601)
		--(axis cs:35.401,9.601)
		--(axis cs:36,9.601)
		--(axis cs:36.001,9.601)
		--(axis cs:36.6,9.601)
		--(axis cs:36.6,9.901)
		--(axis cs:37.2,9.901)
		--(axis cs:37.201,9.901)
		--(axis cs:37.8,9.901)
		--(axis cs:37.801,9.901)
		--(axis cs:38.4,9.901)
		--(axis cs:38.4,10.201)
		--(axis cs:39,10.201)
		--(axis cs:39.001,10.201)
		--(axis cs:39.6,10.201)
		--(axis cs:39.601,10.201)
		--(axis cs:40.2,10.201)
		--(axis cs:40.2,10.501)
		--(axis cs:40.8,10.501)
		--(axis cs:40.801,10.501)
		--(axis cs:41.4,10.501)
		--(axis cs:41.401,10.501)
		--(axis cs:42,10.501)
		--(axis cs:42,10.801)
		--(axis cs:42.6,10.801)
		--(axis cs:42.601,10.801)
		--(axis cs:43.2,10.801)
		--(axis cs:43.201,10.801)
		--(axis cs:43.8,10.801)
		--(axis cs:43.8,11.101)
		--(axis cs:44.4,11.101)
		--(axis cs:44.401,11.101)
		--(axis cs:45,11.101)
		--(axis cs:45.001,11.101)
		--(axis cs:45.6,11.101)
		--(axis cs:45.6,11.401)
		--(axis cs:46.2,11.401)
		--(axis cs:46.201,11.401)
		--(axis cs:46.8,11.401)
		--(axis cs:46.801,11.401)
		--(axis cs:47.4,11.401)
		--(axis cs:47.4,11.701)
		--(axis cs:48,11.701)
		--(axis cs:48.001,11.701)
		--(axis cs:48.6,11.701)
		--(axis cs:48.601,11.701)
		--(axis cs:49.2,11.701)
		--(axis cs:49.2,12.001)
		--(axis cs:49.8,12.001)
		--(axis cs:49.801,12.001)
		--(axis cs:50.4,12.001)
		--(axis cs:50.401,12.001)
		--(axis cs:51,12.001)
		--(axis cs:51.001,12.001)
		--(axis cs:51.6,12.001)
		--(axis cs:51.6,12.301)
		--(axis cs:52.2,12.301)
		--(axis cs:52.201,12.301)
		--(axis cs:52.8,12.301)
		--(axis cs:52.801,12.301)
		--(axis cs:53.4,12.301)
		--(axis cs:53.401,12.301)
		--(axis cs:54,12.301)
		--(axis cs:54,12.601)
		--(axis cs:54.6,12.601)
		--(axis cs:54.601,12.601)
		--(axis cs:55.2,12.601)
		--(axis cs:55.201,12.601)
		--(axis cs:55.8,12.601)
		--(axis cs:55.801,12.601)
		--(axis cs:56.4,12.601)
		--(axis cs:56.4,12.901)
		--(axis cs:57,12.901)
		--(axis cs:57.001,12.901)
		--(axis cs:57.6,12.901)
		--(axis cs:57.601,12.901)
		--(axis cs:58.2,12.901)
		--(axis cs:58.201,12.901)
		--(axis cs:58.8,12.901)
		--(axis cs:58.8,13.201)
		--(axis cs:59.4,13.201)
		--(axis cs:59.401,13.201)
		--(axis cs:60.001,13.201)
		--(axis cs:60.001,12.901)
		--(axis cs:60.001,12.9)
		--(axis cs:60.001,12.601)
		--(axis cs:60.001,12.6)
		--(axis cs:60.001,12.301)
		--(axis cs:60.001,12.3)
		--(axis cs:60.001,12.001)
		--(axis cs:60.001,12)
		--(axis cs:60.001,11.701)
		--(axis cs:60.001,11.7)
		--(axis cs:60.001,11.401)
		--(axis cs:60.001,11.4)
		--(axis cs:60.001,11.101)
		--(axis cs:60.001,11.1)
		--(axis cs:60.001,10.801)
		--(axis cs:60.001,10.8)
		--(axis cs:60.001,10.501)
		--(axis cs:60.001,10.5)
		--(axis cs:60.001,10.201)
		--(axis cs:60.001,10.2)
		--(axis cs:60.001,9.901)
		--(axis cs:60.001,9.9)
		--(axis cs:60.001,9.601)
		--(axis cs:60.001,9.6)
		--(axis cs:60.001,9.301)
		--(axis cs:60.001,9.3)
		--(axis cs:60.001,9.001)
		--(axis cs:60.001,9)
		--(axis cs:60.001,8.701)
		--(axis cs:60.001,8.7)
		--(axis cs:60.001,8.401)
		--(axis cs:60.001,8.4)
		--(axis cs:60.001,8.101)
		--(axis cs:60.001,8.1)
		--(axis cs:60.001,7.801)
		--(axis cs:60.001,7.8)
		--(axis cs:60.001,7.501)
		--(axis cs:60.001,7.5)
		--(axis cs:60.001,7.201)
		--(axis cs:60.001,7.2)
		--(axis cs:60.001,6.901)
		--(axis cs:60.001,6.9)
		--(axis cs:60.001,6.601)
		--(axis cs:60.001,6.6)
		--(axis cs:60.001,6.301)
		--(axis cs:60.001,6.3)
		--(axis cs:60.001,6.001)
		--(axis cs:60.001,6)
		--(axis cs:60.001,5.701)
		--(axis cs:60.001,5.7)
		--(axis cs:60.001,5.401)
		--(axis cs:60.001,5.4)
		--(axis cs:60.001,5.101)
		--(axis cs:60.001,5.1)
		--(axis cs:60.001,4.801)
		--(axis cs:60.001,4.8)
		--(axis cs:60.001,4.501)
		--(axis cs:60.001,4.5)
		--(axis cs:60.001,4.201)
		--(axis cs:60.001,4.2)
		--(axis cs:60.001,3.901)
		--(axis cs:60.001,3.9)
		--(axis cs:60.001,3.601)
		--(axis cs:60.001,3.6)
		--(axis cs:60.001,3.301)
		--(axis cs:60.001,3.3)
		--(axis cs:60.001,3.001)
		--(axis cs:60.001,3)
		--(axis cs:60.001,2.701)
		--(axis cs:60.001,2.7)
		--(axis cs:60.001,2.401)
		--(axis cs:60.001,2.4)
		--(axis cs:60.001,2.101)
		--(axis cs:60.001,2.1)
		--(axis cs:60.001,1.801)
		--(axis cs:60.001,1.8)
		--(axis cs:60.001,1.501)
		--(axis cs:60.001,1.5)
		--(axis cs:60.001,1.201)
		--(axis cs:60.001,1.2)
		--(axis cs:60.001,0.901)
		--(axis cs:60.001,0.9)
		--(axis cs:60.001,0.601)
		--(axis cs:60.001,0.6)
		--(axis cs:60.001,0.301)
		--(axis cs:60.001,0.3)
		--(axis cs:60.001,0)
		--(axis cs:59.401,0)
		--(axis cs:59.4,0)
		--(axis cs:58.801,0)
		--(axis cs:58.8,0)
		--(axis cs:58.201,0)
		--(axis cs:58.2,0)
		--(axis cs:57.601,0)
		--(axis cs:57.6,0)
		--(axis cs:57.001,0)
		--(axis cs:57,0)
		--(axis cs:56.401,0)
		--(axis cs:56.4,0)
		--(axis cs:55.801,0)
		--(axis cs:55.8,0)
		--(axis cs:55.201,0)
		--(axis cs:55.2,0)
		--(axis cs:54.601,0)
		--(axis cs:54.6,0)
		--(axis cs:54.001,0)
		--(axis cs:54,0)
		--(axis cs:53.401,0)
		--(axis cs:53.4,0)
		--(axis cs:52.801,0)
		--(axis cs:52.8,0)
		--(axis cs:52.201,0)
		--(axis cs:52.2,0)
		--(axis cs:51.601,0)
		--(axis cs:51.6,0)
		--(axis cs:51.001,0)
		--(axis cs:51,0)
		--(axis cs:50.401,0)
		--(axis cs:50.4,0)
		--(axis cs:49.801,0)
		--(axis cs:49.8,0)
		--(axis cs:49.201,0)
		--(axis cs:49.2,0)
		--(axis cs:48.601,0)
		--(axis cs:48.6,0)
		--(axis cs:48.001,0)
		--(axis cs:48,0)
		--(axis cs:47.401,0)
		--(axis cs:47.4,0)
		--(axis cs:46.801,0)
		--(axis cs:46.8,0)
		--(axis cs:46.201,0)
		--(axis cs:46.2,0)
		--(axis cs:45.601,0)
		--(axis cs:45.6,0)
		--(axis cs:45.001,0)
		--(axis cs:45,0)
		--(axis cs:44.401,0)
		--(axis cs:44.4,0)
		--(axis cs:43.801,0)
		--(axis cs:43.8,0)
		--(axis cs:43.201,0)
		--(axis cs:43.2,0)
		--(axis cs:42.601,0)
		--(axis cs:42.6,0)
		--(axis cs:42.001,0)
		--(axis cs:42,0)
		--(axis cs:41.401,0)
		--(axis cs:41.4,0)
		--(axis cs:40.801,0)
		--(axis cs:40.8,0)
		--(axis cs:40.201,0)
		--(axis cs:40.2,0)
		--(axis cs:39.601,0)
		--(axis cs:39.6,0)
		--(axis cs:39.001,0)
		--(axis cs:39,0)
		--(axis cs:38.401,0)
		--(axis cs:38.4,0)
		--(axis cs:37.801,0)
		--(axis cs:37.8,0)
		--(axis cs:37.201,0)
		--(axis cs:37.2,0)
		--(axis cs:36.601,0)
		--(axis cs:36.6,0)
		--(axis cs:36.001,0)
		--(axis cs:36,0)
		--(axis cs:35.401,0)
		--(axis cs:35.4,0)
		--(axis cs:34.801,0)
		--(axis cs:34.8,0)
		--(axis cs:34.201,0)
		--(axis cs:34.2,0)
		--(axis cs:33.601,0)
		--(axis cs:33.6,0)
		--(axis cs:33.001,0)
		--(axis cs:33,0)
		--(axis cs:32.401,0)
		--(axis cs:32.4,0)
		--(axis cs:31.801,0)
		--(axis cs:31.8,0)
		--(axis cs:31.201,0)
		--(axis cs:31.2,0)
		--(axis cs:30.601,0)
		--(axis cs:30.6,0)
		--(axis cs:30.001,0)
		--(axis cs:30,0)
		--(axis cs:29.401,0)
		--(axis cs:29.4,0)
		--(axis cs:28.801,0)
		--(axis cs:28.8,0)
		--(axis cs:28.201,0)
		--(axis cs:28.2,0)
		--(axis cs:27.601,0)
		--(axis cs:27.6,0)
		--(axis cs:27.001,0)
		--(axis cs:27,0)
		--(axis cs:26.401,0)
		--(axis cs:26.4,0)
		--(axis cs:25.801,0)
		--(axis cs:25.8,0)
		--(axis cs:25.201,0)
		--(axis cs:25.2,0)
		--(axis cs:24.601,0)
		--(axis cs:24.6,0)
		--(axis cs:24.001,0)
		--(axis cs:24,0)
		--(axis cs:23.401,0)
		--(axis cs:23.4,0)
		--(axis cs:22.801,0)
		--(axis cs:22.8,0)
		--(axis cs:22.201,0)
		--(axis cs:22.2,0)
		--(axis cs:21.601,0)
		--(axis cs:21.6,0)
		--(axis cs:21.001,0)
		--(axis cs:21,0)
		--(axis cs:20.401,0)
		--(axis cs:20.4,0)
		--(axis cs:19.801,0)
		--(axis cs:19.8,0)
		--(axis cs:19.201,0)
		--(axis cs:19.2,0)
		--(axis cs:18.601,0)
		--(axis cs:18.6,0)
		--(axis cs:18.001,0)
		--(axis cs:18,0)
		--(axis cs:17.401,0)
		--(axis cs:17.4,0)
		--(axis cs:16.801,0)
		--(axis cs:16.8,0)
		--(axis cs:16.201,0)
		--(axis cs:16.2,0)
		--(axis cs:15.601,0)
		--(axis cs:15.6,0)
		--(axis cs:15.001,0)
		--(axis cs:15,0)
		--(axis cs:14.401,0)
		--(axis cs:14.4,0)
		--(axis cs:13.801,0)
		--(axis cs:13.8,0)
		--(axis cs:13.201,0)
		--(axis cs:13.2,0)
		--(axis cs:12.601,0)
		--(axis cs:12.6,0)
		--(axis cs:12.001,0)
		--(axis cs:12,0)
		--(axis cs:11.401,0)
		--(axis cs:11.4,0)
		--(axis cs:10.801,0)
		--(axis cs:10.8,0)
		--(axis cs:10.201,0)
		--(axis cs:10.2,0)
		--(axis cs:9.601,0)
		--(axis cs:9.6,0)
		--(axis cs:9.001,0)
		--(axis cs:9,0)
		--(axis cs:8.401,0)
		--(axis cs:8.4,0)
		--(axis cs:7.801,0)
		--(axis cs:7.8,0)
		--(axis cs:7.201,0)
		--(axis cs:7.2,0)
		--(axis cs:6.601,0)
		--(axis cs:6.6,0)
		--(axis cs:6.001,0)
		--(axis cs:6,0)
		--cycle;
		\path [draw=none, fill=blue]
		(axis cs:6.6,1.201)
		--(axis cs:6.601,1.201)
		--(axis cs:7.2,1.201)
		--(axis cs:7.201,1.201)
		--(axis cs:7.8,1.201)
		--(axis cs:7.801,1.201)
		--(axis cs:8.4,1.201)
		--(axis cs:8.401,1.201)
		--(axis cs:9,1.201)
		--(axis cs:9.001,1.201)
		--(axis cs:9.6,1.201)
		--(axis cs:9.601,1.201)
		--(axis cs:10.2,1.201)
		--(axis cs:10.201,1.201)
		--(axis cs:10.8,1.201)
		--(axis cs:10.801,1.201)
		--(axis cs:11.4,1.201)
		--(axis cs:11.401,1.201)
		--(axis cs:12,1.201)
		--(axis cs:12.001,1.201)
		--(axis cs:12.6,1.201)
		--(axis cs:12.601,1.201)
		--(axis cs:13.2,1.201)
		--(axis cs:13.201,1.201)
		--(axis cs:13.8,1.201)
		--(axis cs:13.801,1.201)
		--(axis cs:14.4,1.201)
		--(axis cs:14.401,1.201)
		--(axis cs:15,1.201)
		--(axis cs:15.001,1.201)
		--(axis cs:15.6,1.201)
		--(axis cs:15.601,1.201)
		--(axis cs:16.2,1.201)
		--(axis cs:16.201,1.201)
		--(axis cs:16.8,1.201)
		--(axis cs:16.801,1.201)
		--(axis cs:17.4,1.201)
		--(axis cs:17.401,1.201)
		--(axis cs:18,1.201)
		--(axis cs:18.001,1.201)
		--(axis cs:18.6,1.201)
		--(axis cs:18.601,1.201)
		--(axis cs:19.2,1.201)
		--(axis cs:19.201,1.201)
		--(axis cs:19.8,1.201)
		--(axis cs:19.801,1.201)
		--(axis cs:20.4,1.201)
		--(axis cs:20.401,1.201)
		--(axis cs:21,1.201)
		--(axis cs:21.001,1.201)
		--(axis cs:21.6,1.201)
		--(axis cs:21.601,1.201)
		--(axis cs:22.2,1.201)
		--(axis cs:22.201,1.201)
		--(axis cs:22.8,1.201)
		--(axis cs:22.801,1.201)
		--(axis cs:23.4,1.201)
		--(axis cs:23.401,1.201)
		--(axis cs:24,1.201)
		--(axis cs:24.001,1.201)
		--(axis cs:24.6,1.201)
		--(axis cs:24.601,1.201)
		--(axis cs:25.2,1.201)
		--(axis cs:25.201,1.201)
		--(axis cs:25.8,1.201)
		--(axis cs:25.801,1.201)
		--(axis cs:26.4,1.201)
		--(axis cs:26.401,1.201)
		--(axis cs:27,1.201)
		--(axis cs:27.001,1.201)
		--(axis cs:27.6,1.201)
		--(axis cs:27.601,1.201)
		--(axis cs:28.2,1.201)
		--(axis cs:28.201,1.201)
		--(axis cs:28.8,1.201)
		--(axis cs:28.801,1.201)
		--(axis cs:29.4,1.201)
		--(axis cs:29.401,1.201)
		--(axis cs:30,1.201)
		--(axis cs:30.001,1.201)
		--(axis cs:30.6,1.201)
		--(axis cs:30.601,1.201)
		--(axis cs:31.2,1.201)
		--(axis cs:31.201,1.201)
		--(axis cs:31.8,1.201)
		--(axis cs:31.801,1.201)
		--(axis cs:32.4,1.201)
		--(axis cs:32.401,1.201)
		--(axis cs:33,1.201)
		--(axis cs:33.001,1.201)
		--(axis cs:33.6,1.201)
		--(axis cs:33.601,1.201)
		--(axis cs:34.2,1.201)
		--(axis cs:34.201,1.201)
		--(axis cs:34.8,1.201)
		--(axis cs:34.801,1.201)
		--(axis cs:35.4,1.201)
		--(axis cs:35.401,1.201)
		--(axis cs:36,1.201)
		--(axis cs:36.001,1.201)
		--(axis cs:36.6,1.201)
		--(axis cs:36.601,1.201)
		--(axis cs:37.2,1.201)
		--(axis cs:37.201,1.201)
		--(axis cs:37.8,1.201)
		--(axis cs:37.801,1.201)
		--(axis cs:38.4,1.201)
		--(axis cs:38.401,1.201)
		--(axis cs:39,1.201)
		--(axis cs:39.001,1.201)
		--(axis cs:39.6,1.201)
		--(axis cs:39.601,1.201)
		--(axis cs:40.2,1.201)
		--(axis cs:40.201,1.201)
		--(axis cs:40.8,1.201)
		--(axis cs:40.801,1.201)
		--(axis cs:41.4,1.201)
		--(axis cs:41.401,1.201)
		--(axis cs:42,1.201)
		--(axis cs:42.001,1.201)
		--(axis cs:42.6,1.201)
		--(axis cs:42.601,1.201)
		--(axis cs:43.2,1.201)
		--(axis cs:43.201,1.201)
		--(axis cs:43.8,1.201)
		--(axis cs:43.801,1.201)
		--(axis cs:44.4,1.201)
		--(axis cs:44.401,1.201)
		--(axis cs:45,1.201)
		--(axis cs:45.001,1.201)
		--(axis cs:45.6,1.201)
		--(axis cs:45.601,1.201)
		--(axis cs:46.2,1.201)
		--(axis cs:46.201,1.201)
		--(axis cs:46.8,1.201)
		--(axis cs:46.801,1.201)
		--(axis cs:47.4,1.201)
		--(axis cs:47.401,1.201)
		--(axis cs:48,1.201)
		--(axis cs:48.001,1.201)
		--(axis cs:48.6,1.201)
		--(axis cs:48.601,1.201)
		--(axis cs:49.2,1.201)
		--(axis cs:49.201,1.201)
		--(axis cs:49.8,1.201)
		--(axis cs:49.801,1.201)
		--(axis cs:50.4,1.201)
		--(axis cs:50.401,1.201)
		--(axis cs:51,1.201)
		--(axis cs:51.001,1.201)
		--(axis cs:51.6,1.201)
		--(axis cs:51.601,1.201)
		--(axis cs:52.2,1.201)
		--(axis cs:52.201,1.201)
		--(axis cs:52.8,1.201)
		--(axis cs:52.801,1.201)
		--(axis cs:53.4,1.201)
		--(axis cs:53.401,1.201)
		--(axis cs:54,1.201)
		--(axis cs:54.001,1.201)
		--(axis cs:54.6,1.201)
		--(axis cs:54.601,1.201)
		--(axis cs:55.2,1.201)
		--(axis cs:55.201,1.201)
		--(axis cs:55.8,1.201)
		--(axis cs:55.801,1.201)
		--(axis cs:56.4,1.201)
		--(axis cs:56.401,1.201)
		--(axis cs:57,1.201)
		--(axis cs:57.001,1.201)
		--(axis cs:57.6,1.201)
		--(axis cs:57.601,1.201)
		--(axis cs:58.2,1.201)
		--(axis cs:58.201,1.201)
		--(axis cs:58.8,1.201)
		--(axis cs:58.801,1.201)
		--(axis cs:59.4,1.201)
		--(axis cs:59.401,1.201)
		--(axis cs:60.001,1.201)
		--(axis cs:60.001,0.901)
		--(axis cs:60.001,0.9)
		--(axis cs:60.001,0.601)
		--(axis cs:60.001,0.6)
		--(axis cs:60.001,0.301)
		--(axis cs:60.001,0.3)
		--(axis cs:60.001,0)
		--(axis cs:59.401,0)
		--(axis cs:59.4,0)
		--(axis cs:58.801,0)
		--(axis cs:58.8,0)
		--(axis cs:58.201,0)
		--(axis cs:58.2,0)
		--(axis cs:57.601,0)
		--(axis cs:57.6,0)
		--(axis cs:57.001,0)
		--(axis cs:57,0)
		--(axis cs:56.401,0)
		--(axis cs:56.4,0)
		--(axis cs:55.801,0)
		--(axis cs:55.8,0)
		--(axis cs:55.201,0)
		--(axis cs:55.2,0)
		--(axis cs:54.601,0)
		--(axis cs:54.6,0)
		--(axis cs:54.001,0)
		--(axis cs:54,0)
		--(axis cs:53.401,0)
		--(axis cs:53.4,0)
		--(axis cs:52.801,0)
		--(axis cs:52.8,0)
		--(axis cs:52.201,0)
		--(axis cs:52.2,0)
		--(axis cs:51.601,0)
		--(axis cs:51.6,0)
		--(axis cs:51.001,0)
		--(axis cs:51,0)
		--(axis cs:50.401,0)
		--(axis cs:50.4,0)
		--(axis cs:49.801,0)
		--(axis cs:49.8,0)
		--(axis cs:49.201,0)
		--(axis cs:49.2,0)
		--(axis cs:48.601,0)
		--(axis cs:48.6,0)
		--(axis cs:48.001,0)
		--(axis cs:48,0)
		--(axis cs:47.401,0)
		--(axis cs:47.4,0)
		--(axis cs:46.801,0)
		--(axis cs:46.8,0)
		--(axis cs:46.201,0)
		--(axis cs:46.2,0)
		--(axis cs:45.601,0)
		--(axis cs:45.6,0)
		--(axis cs:45.001,0)
		--(axis cs:45,0)
		--(axis cs:44.401,0)
		--(axis cs:44.4,0)
		--(axis cs:43.801,0)
		--(axis cs:43.8,0)
		--(axis cs:43.201,0)
		--(axis cs:43.2,0)
		--(axis cs:42.601,0)
		--(axis cs:42.6,0)
		--(axis cs:42.001,0)
		--(axis cs:42,0)
		--(axis cs:41.401,0)
		--(axis cs:41.4,0)
		--(axis cs:40.801,0)
		--(axis cs:40.8,0)
		--(axis cs:40.201,0)
		--(axis cs:40.2,0)
		--(axis cs:39.601,0)
		--(axis cs:39.6,0)
		--(axis cs:39.001,0)
		--(axis cs:39,0)
		--(axis cs:38.401,0)
		--(axis cs:38.4,0)
		--(axis cs:37.801,0)
		--(axis cs:37.8,0)
		--(axis cs:37.201,0)
		--(axis cs:37.2,0)
		--(axis cs:36.601,0)
		--(axis cs:36.6,0)
		--(axis cs:36.001,0)
		--(axis cs:36,0)
		--(axis cs:35.401,0)
		--(axis cs:35.4,0)
		--(axis cs:34.801,0)
		--(axis cs:34.8,0)
		--(axis cs:34.201,0)
		--(axis cs:34.2,0)
		--(axis cs:33.601,0)
		--(axis cs:33.6,0)
		--(axis cs:33.001,0)
		--(axis cs:33,0)
		--(axis cs:32.401,0)
		--(axis cs:32.4,0)
		--(axis cs:31.801,0)
		--(axis cs:31.8,0)
		--(axis cs:31.201,0)
		--(axis cs:31.2,0)
		--(axis cs:30.601,0)
		--(axis cs:30.6,0)
		--(axis cs:30.001,0)
		--(axis cs:30,0)
		--(axis cs:29.401,0)
		--(axis cs:29.4,0)
		--(axis cs:28.801,0)
		--(axis cs:28.8,0)
		--(axis cs:28.201,0)
		--(axis cs:28.2,0)
		--(axis cs:27.601,0)
		--(axis cs:27.6,0)
		--(axis cs:27.001,0)
		--(axis cs:27,0)
		--(axis cs:26.401,0)
		--(axis cs:26.4,0)
		--(axis cs:25.801,0)
		--(axis cs:25.8,0)
		--(axis cs:25.201,0)
		--(axis cs:25.2,0)
		--(axis cs:24.601,0)
		--(axis cs:24.6,0)
		--(axis cs:24.001,0)
		--(axis cs:24,0)
		--(axis cs:23.401,0)
		--(axis cs:23.4,0)
		--(axis cs:22.801,0)
		--(axis cs:22.8,0)
		--(axis cs:22.201,0)
		--(axis cs:22.2,0)
		--(axis cs:21.601,0)
		--(axis cs:21.6,0)
		--(axis cs:21.001,0)
		--(axis cs:21,0)
		--(axis cs:20.401,0)
		--(axis cs:20.4,0)
		--(axis cs:19.801,0)
		--(axis cs:19.8,0)
		--(axis cs:19.201,0)
		--(axis cs:19.2,0)
		--(axis cs:18.601,0)
		--(axis cs:18.6,0)
		--(axis cs:18.001,0)
		--(axis cs:18,0)
		--(axis cs:17.401,0)
		--(axis cs:17.4,0)
		--(axis cs:16.801,0)
		--(axis cs:16.8,0)
		--(axis cs:16.201,0)
		--(axis cs:16.2,0)
		--(axis cs:15.601,0)
		--(axis cs:15.6,0)
		--(axis cs:15.001,0)
		--(axis cs:15,0)
		--(axis cs:14.401,0)
		--(axis cs:14.4,0)
		--(axis cs:13.801,0)
		--(axis cs:13.8,0)
		--(axis cs:13.201,0)
		--(axis cs:13.2,0)
		--(axis cs:12.601,0)
		--(axis cs:12.6,0)
		--(axis cs:12.001,0)
		--(axis cs:12,0)
		--(axis cs:11.401,0)
		--(axis cs:11.4,0)
		--(axis cs:10.801,0)
		--(axis cs:10.8,0)
		--(axis cs:10.201,0)
		--(axis cs:10.2,0)
		--(axis cs:9.601,0)
		--(axis cs:9.6,0)
		--(axis cs:9.001,0)
		--(axis cs:9,0)
		--(axis cs:8.401,0)
		--(axis cs:8.4,0)
		--(axis cs:7.801,0)
		--(axis cs:7.8,0)
		--(axis cs:7.201,0)
		--(axis cs:7.2,0)
		--(axis cs:6.601,0)
		--(axis cs:6.6,0)
		--(axis cs:6.001,0)
		--(axis cs:6,0)
		--(axis cs:5.401,0)
		--(axis cs:5.4,0)
		--(axis cs:4.801,0)
		--(axis cs:4.8,0)
		--(axis cs:4.201,0)
		--(axis cs:4.2,0)
		--(axis cs:3.601,0)
		--(axis cs:3.6,0)
		--(axis cs:3.001,0)
		--(axis cs:3,0)
		--(axis cs:2.401,0)
		--(axis cs:2.4,0)
		--(axis cs:1.801,0)
		--(axis cs:1.8,0)
		--(axis cs:1.201,0)
		--(axis cs:1.2,0)
		--(axis cs:0.6,0)
		--(axis cs:0.6,0.301)
		--(axis cs:1.2,0.301)
		--(axis cs:1.201,0.301)
		--(axis cs:1.8,0.301)
		--(axis cs:1.801,0.301)
		--(axis cs:2.4,0.301)
		--(axis cs:2.4,0.601)
		--(axis cs:3,0.601)
		--(axis cs:3,0.901)
		--(axis cs:3.6,0.901)
		--(axis cs:3.6,1.201)
		--(axis cs:4.2,1.201)
		--(axis cs:4.201,1.201)
		--(axis cs:4.8,1.201)
		--(axis cs:4.801,1.201)
		--(axis cs:5.4,1.201)
		--(axis cs:5.401,1.201)
		--(axis cs:6,1.201)
		--(axis cs:6.001,1.201)
		--cycle;
		\node[align=center,font=\scriptsize] at (axis cs: 15,23.8) {Unsafe (sim)};
		\node[align=center,font=\scriptsize] at (axis cs: 36,15.8) {Inconclusive};
		\node[align=center,font=\scriptsize] at (axis cs: 40,5) {Safe ($3$-step)};
	\end{axis}

\end{tikzpicture}

%% file: conclusion.tex
\section{Conclusions}
Analysis of image-based NNCS typically involves running lots of tests.
By leveraging generative models to approximate the perception system, 
the surrogate verification approach offers an alternative that can utilize computational set-based analysis methods.
%
Scalability is the primary bottleneck in this line of work and the main problem addressed in this paper.
We identified underlying causes of overapproximation error with the existing baseline method and
proposed two strategies--composition and unrolling--to overcome these problems.
Our evaluation demonstrates the advantages of our approach over the baseline.
In terms of accuracy, 
our method reduces the converged reachable set by $175\%$ on the aircraft taxiing system compared to the baseline.
%
In terms of scalability, the transformer variant of the cGAN in emergency braking case study has 24$\times$ more output pixels than the prior work's case study. 

%
Despite improvements, scalability is not yet solved, as 
state-of-the-art generative models complex and continue to grow in size.
Increasing the ranges and number of latent variables, generating higher-resolution images, considering systems with more state variables and incorporating recent video GAN techniques~\cite{tulyakov2018mocogan} will all require further improvements to scalability.


%% file: acknowledgments.tex
\section{Acknowledgments}
This material is based upon work supported by the Air Force Office of Scientific Research and the Office of Naval Research under award numbers FA9550-19-1-0288, FA9550-21-1-0121, FA9550-23-1-0066 and N00014-22-1-2156, and the National Science Foundation under Award No. 2237229. Any opinions, findings, and conclusions or recommendations expressed in this material are those of the author(s) and do not necessarily reflect the views of the United States Air Force or the United States Navy. 

%% file: appendix.tex
\makeatletter

\renewcommand{\thesection}{\Alph{section}}
\setcounter{secnumdepth}{2}
\makeatother

\section{Appendix for Methodology}
\label{appendix:methodology}
\subsection{Forward Reachability Algorithm}
\textbf{Completeness Justification:} The forward reachability algorithm iteratively computes the reachable set using the functions $\FReach$ and the abstraction process $\alpha$. 
The completeness of the algorithm is ensured by the properties of these components.
The abstraction process $\alpha$ identifies the set of cells $\mathcal{C}$ that overlap with the reachable region $\mathcal{R}$, providing an overapproximation of the exact reachable set $R$. 
This guarantees that all reachable states are included, ensuring the abstraction process captures the entire reachable region.
The computation of reachable sets through $\FReach$ is complete, as it relies on two composition options whose completeness is guaranteed by their underlying algorithms. The first option uses neural network verification tools with proven completeness guarantees, while the second leverages reachability algorithms from hybrid systems, which are also formally complete.
\newline
\textbf{Convergent Justification:} 
The forward reachability algorithm is guaranteed to converge, as it terminates either when the reachable set stabilizes or when the predefined maximum time step $k_\text{max}$ is reached.

%
%
%
%

\subsection{Backward Reachability Algorithm}
The algorithm utilizes the backward reachability analysis to determine a set of cells $\mathcal{A}$ within the state space that may potentially lead to unsafe system states,
and the set of cells that can be guaranteed to be safe is essentially the complement of set $\mathcal{A}$.
To streamline the process, the function of computing the reachable set (\FReach) and the abstraction process ($\alpha$) can be combined into a single function \FReachCell. This function calculates the forward reachable cells starting from a single cell or a set of cells. 
Given that the input space is discretized into a finite number of cells, 
evaluating the forward reachable cells for each individual cell enables the construction of a reverse mapping \FBackReachCell, which can identify all cells that can reach a target cell or a set of target cells.

\begin{algorithm}[!ht]
	\SetAlgoLined
	\DontPrintSemicolon
	\Fn{\FBackMain}{
		\KwInput{$\mathcal{U}$, unsafe region}
		\KwInput{$m$, unrolling steps}
		\KwInput{$\mathcal{H}$, rectangular cells defined within the state space}
		
		\tcc{initialize the possible unsafe set}
		$\mathcal{A} : = \mathcal{U} \cap \mathcal{H}$\;
		\ForEach{$c \in \mathcal{H}$}
		{
			\For {$i=1$ \KwTo $m-1$}
			{
				$\mathcal{C} := \FReachCell_i(c)$\;
				\If{$\mathcal{C} \cap \mathcal{U} \neq \emptyset$}
				{$\mathcal{A} := \mathcal{A} \cup \{c\}$}
			}
		}
		
		\tcc{Compute all possible unsafe cells using $m$-step backward reachability}
		$\mathcal{N} := \mathcal{A}$\;
		\While{$\mathcal{N} \neq \emptyset$}
		{
			$\mathcal{N}^\prime := \emptyset$\;
			\ForEach{$c \in \mathcal{N}$}
			{$\mathcal{C} := \FBackReachCell_m(c)$\;
				$\mathcal{N}^\prime := \mathcal{N}^\prime \cup \mathcal{C}$ \;				
			}
			$\mathcal{N} := \mathcal{A} \cap \mathcal{N}^\prime$\;
			$\mathcal{A} := \mathcal{A} \cup \mathcal{N}^\prime$
		}
		
		\KwRet{$\mathcal{A}$}
	}
	\caption{Proposed backward reachability algorithm.}
	\label{alg:back_reach_analysis}
\end{algorithm}

The algorithm for backward reachability analysis
begins by initializing the set of potentially unsafe cells $\mathcal{A}$. This is done by considering the cells themselves as unsafe (line 2) or identifying cells that can reach an unsafe state through propagation using any one of the methods ranging from  $1$-step to ($m-1$)-step approaches (lines 3-10). 
For each cell $c$ that is newly added to $\mathcal{A}$, the algorithm computes all cells that could reach $c$ using an $m$-step backward reachability analysis. These newly identified cells are considered potentially unsafe and are added to the set $\mathcal{A}$. This iteration continues until no new cells are added to $\mathcal{A}$, and the algorithm returns the set of possible unsafe cells $\mathcal{A}$ within the state space.
\newline
\textbf{Completeness Justification:} If a cell is unsafe, 
it must reach an unsafe region at some step $k$ in the future.
Consider $k//m = a$ and $t\%m = b$.
This cell must be included in the set computed using the following procedure defined within the algorithm. 
This set is obtained by first finding the set of cells that can reach an unsafe state using a $b$-step method (initialization procedure in the algorithm) and then back-propagating this set $a$ times using an $m$-step method (iteration procedure in the algorithm).
\newline 
\textbf{Convergent Justification:} 
Since the state space is quantized into a finite number of cells, it is impossible 
to add new cells to the set $\mathcal{A}$ infinitely.

\section{Appendix for Autonomous Aircraft Taxiing System}
\label{appendix:aats}
\subsection{Network Architecture and Generated Images}
In the prior work with the baseline method~\cite{katz2022verification}, several simplifications were made to make the verification of the surrogate system tractable.
First, the input images, originally $200\times360$ color images, were downsampled to $8\times16$ grayscale images.
Second, the cGAN was initially trained using a deep convolutional GAN (DCGAN), 
but was then replaced by a smaller feed-forward neural network that emulates the image generation process of the DCGAN, in order to simplify analysis by the neural network verification tool.
The detail architecture of the unified network is listed in Table~\ref{tab:network_architectures_taxi}.
Also, two pairs of real images and the generated images using the network are shown in Fig.~\ref{fig:gan_images_aats}.

\begin{table}[!h]
	\centering
	\normalsize
	\caption{Architectures of neural networks in autonomous aircraft taxiing system.}
	\label{tab:network_architectures_taxi}
		\centering
		\begin{tabular}{c}
			\hhline{=}
			$z \in \mathbb{R}^2; p \in \mathbb{R}; \theta \in \mathbb{R}$
			\\
			\hline
			$\mathsf{Concat} (p, \theta, z) \in \mathbb{R}^4 $
			\\
			\hline
			$\mathsf{Dense} \rightarrow 256$, $\mathsf{ReLU}$
			\\
			\hline
			$\mathsf{Dense} \rightarrow 256$, $\mathsf{ReLU}$
			\\
			\hline
			$\mathsf{Dense} \rightarrow 256$, $\mathsf{ReLU}$
			\\
			\hline
			$\mathsf{Dense} \rightarrow 256$, $\mathsf{ReLU}$
			\\
			\hline
			$\mathsf{Dense} \rightarrow 16$, $\mathsf{ReLU}$
			\\
			\hline
			$\mathsf{Dense} \rightarrow 8$, $\mathsf{ReLU}$
			\\
			\hline
			$\mathsf{Dense} \rightarrow 8$, $\mathsf{ReLU}$
			\\
			\hline
			$\mathsf{Dense} \rightarrow 2$
			\\
			\hhline{=}
		\end{tabular}
\end{table}

\begin{figure}[!t]
	\centering
	\tikzsetnextfilename{real_images_generated_images_aircraft}
	\input{figures/real_images_generated_images_aircraft}
	\caption{Real images and generated images for the taxiing system, the case study considered by the baseline method. }
	\label{fig:gan_images_aats}
\end{figure}


\subsection{Linearization Technique for the Dynamics}

To account for the nonlinear dynamics of aircraft taxiing system,
we employ a conservative linearization technique based on Taylor expansion~\cite{althoff2008reachability}.
This technique approximates the nonlinear dynamics with a \nth{1} order Taylor series and its \nth{2} order remainder:
\begin{equation*}
		x_{k+1}^i \in 	\overbrace{f^i(z_k^\ast) + \underbrace{\frac{\partial f^i (z)}{\partial z}\bigg|_{z_k^\ast}}_{A_{(i, \cdot)}} (z_k - z_k^\ast )}^{\text{\nth{1} order Taylor expansion}}\ +\ L_i,
\end{equation*}
where $x^T = [p, \theta]^T$ is the state vector, $z^T = [p, \theta, \phi]^T$ combines the states and input, 
and $z^\ast$ is the expansion point, typically located at the center of the star set.
Specifically, $A = \begin{bmatrix}
		1 & v\Delta t \cos\theta & 0 \\
		0 & 1 & \frac{v}{L}\Delta t\tan^2\phi
	\end{bmatrix}$,  
	and the Lagrange remainders for the two state variables can be  separately overapproximated using interval arithmetic~\cite{jaulin2001interval} as:
	\begin{equation*}
		\begin{split}
				& \begin{split}
						L_1\subseteq \frac{1}{2}v\Delta t[
								&\min(-\sin \theta \cdot (\theta - \theta^\ast)^2), \\
						        &\max(-\sin \theta \cdot (\theta - \theta^\ast)^2)],
					\end{split}\\
				& \begin{split}
						L_2\subseteq \frac{v}{L}\Delta t[
							&\min(\tan \phi \cdot (\tan^2\phi+1) \cdot (\phi- \phi^\ast)^2), \\
							&\max(\tan \phi \cdot (\tan^2\phi+1) \cdot (\phi- \phi^\ast)^2)].
					\end{split}	
			\end{split}
	\end{equation*}

\subsection{Additional Experiments}
\begin{figure*}[!t]
	\centering
	\subfloat{\includegraphics{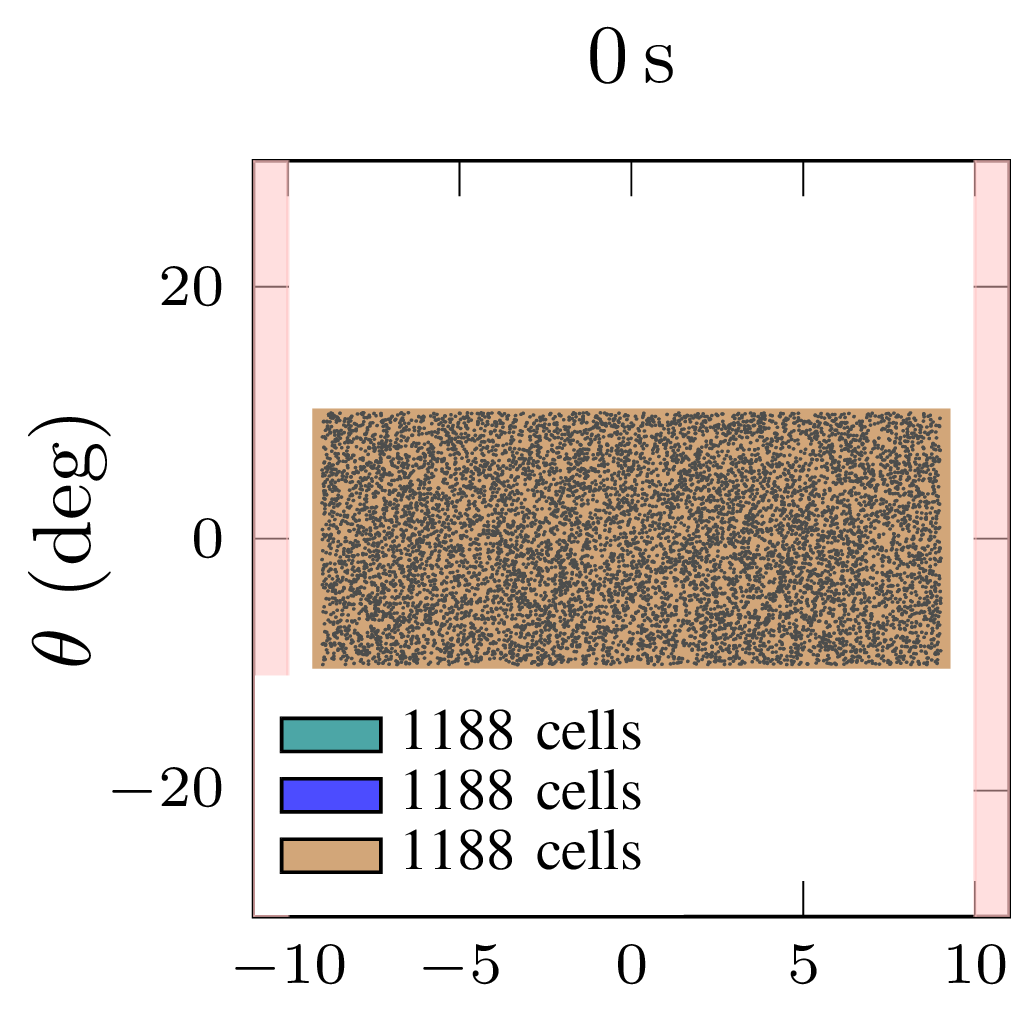}}
	\hspace{-1.0em}
	\subfloat{\includegraphics{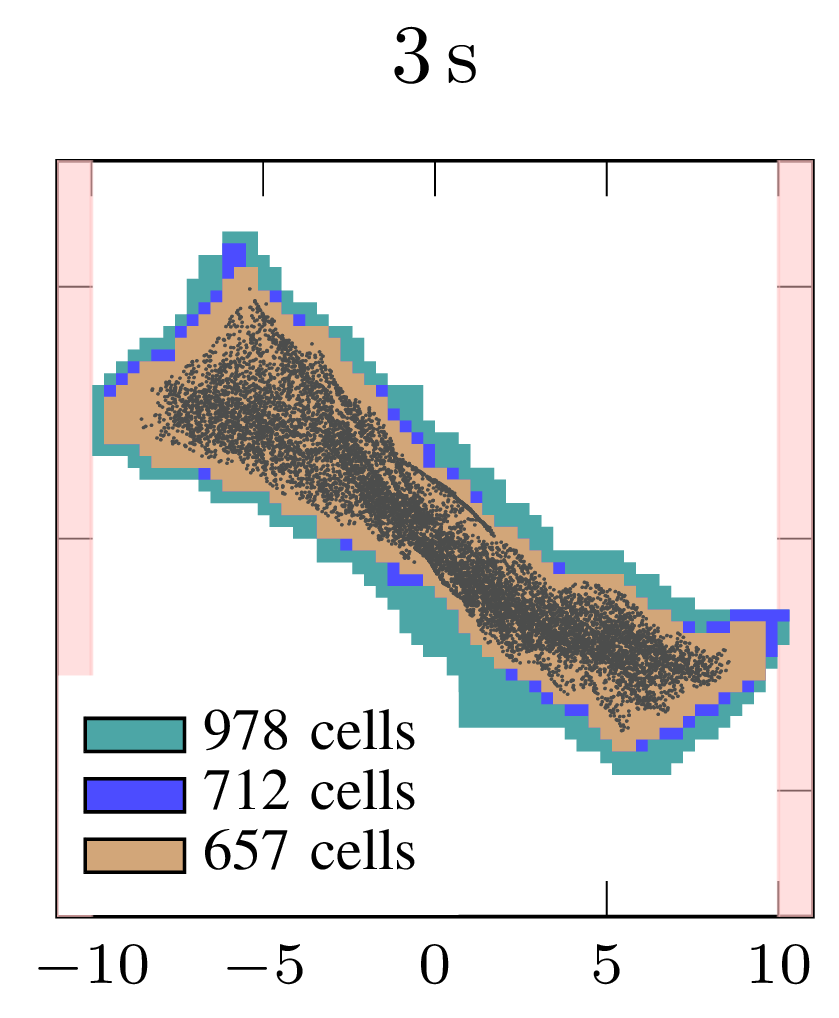}}
	\hspace{-1.0em}
	\subfloat{\includegraphics{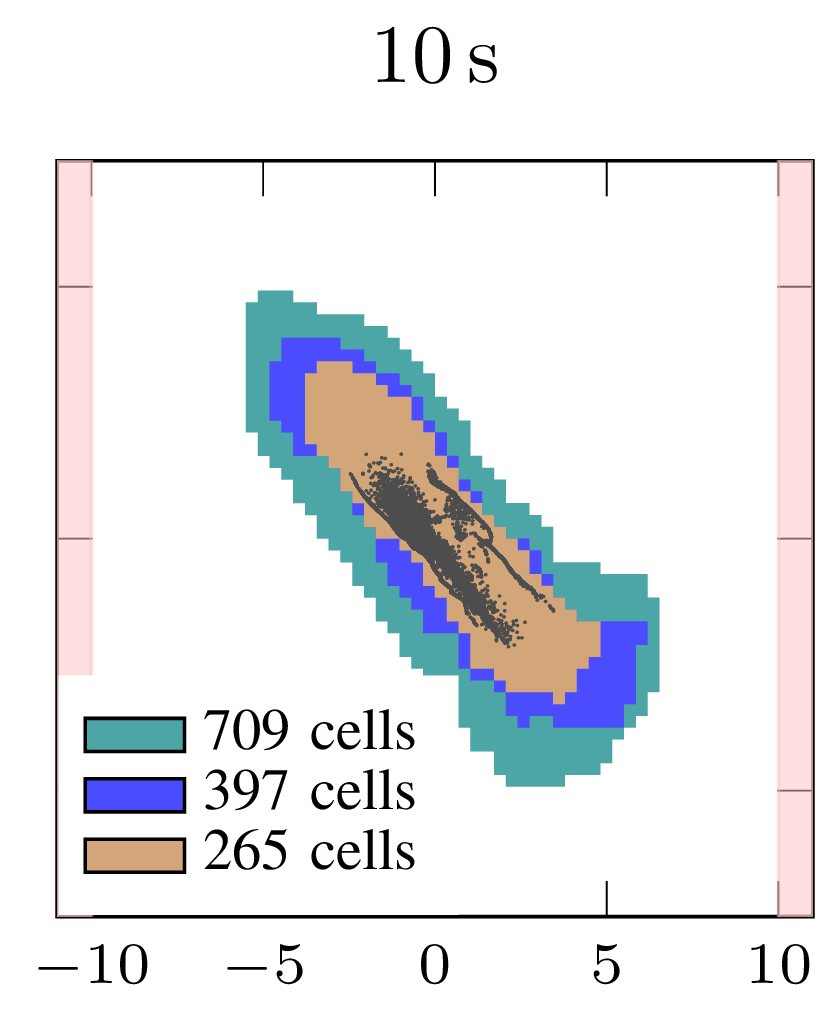}}
	\hspace{-1.0em}
	\subfloat{\includegraphics{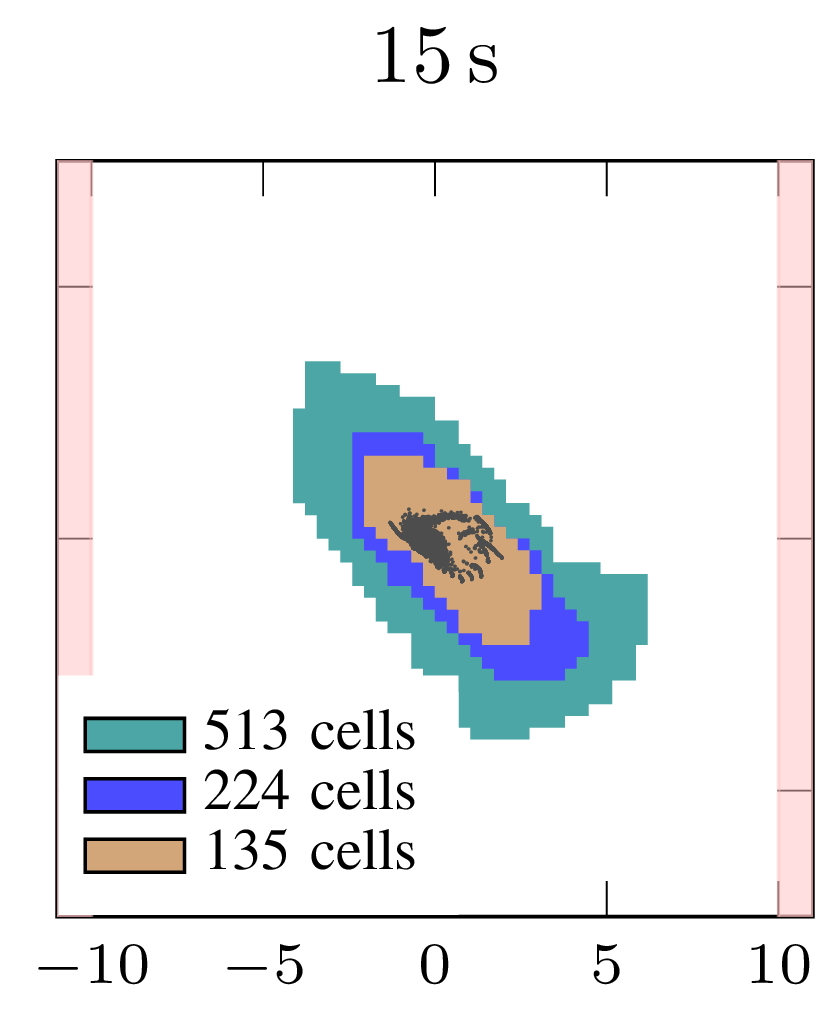}}
	\hspace{-1.0em}
	\subfloat{\includegraphics{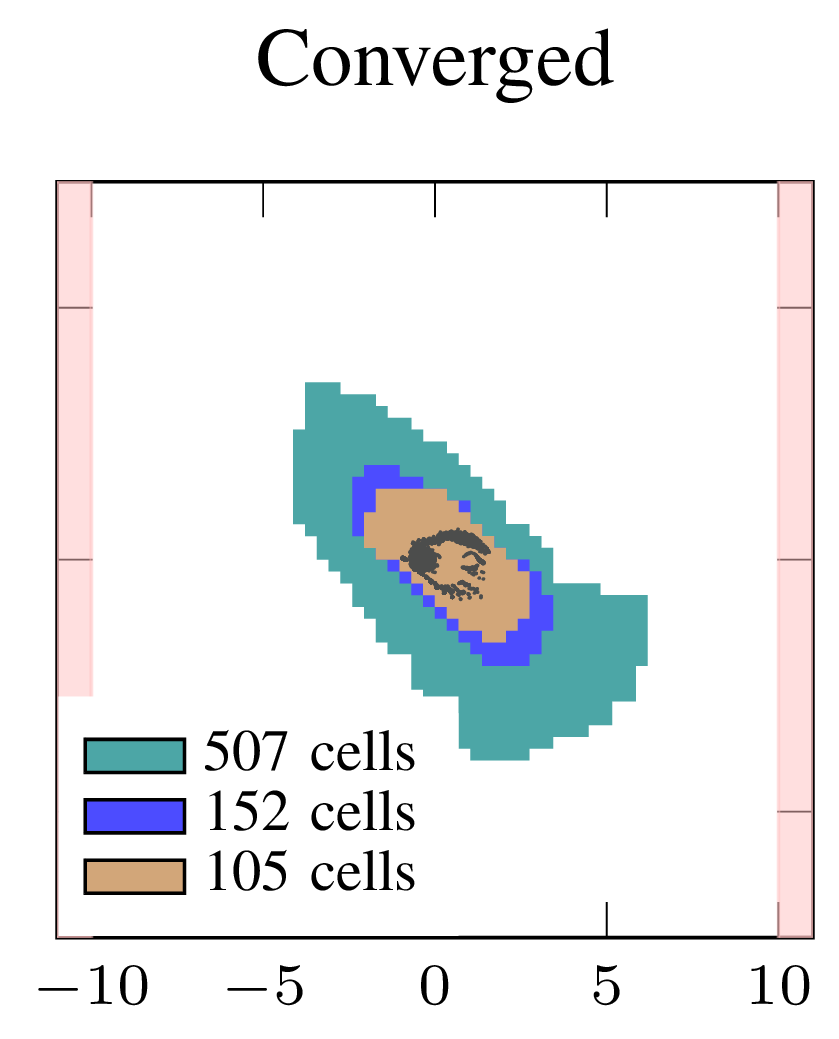}}
	\hspace{-1.0em}
	\subfloat{\includegraphics{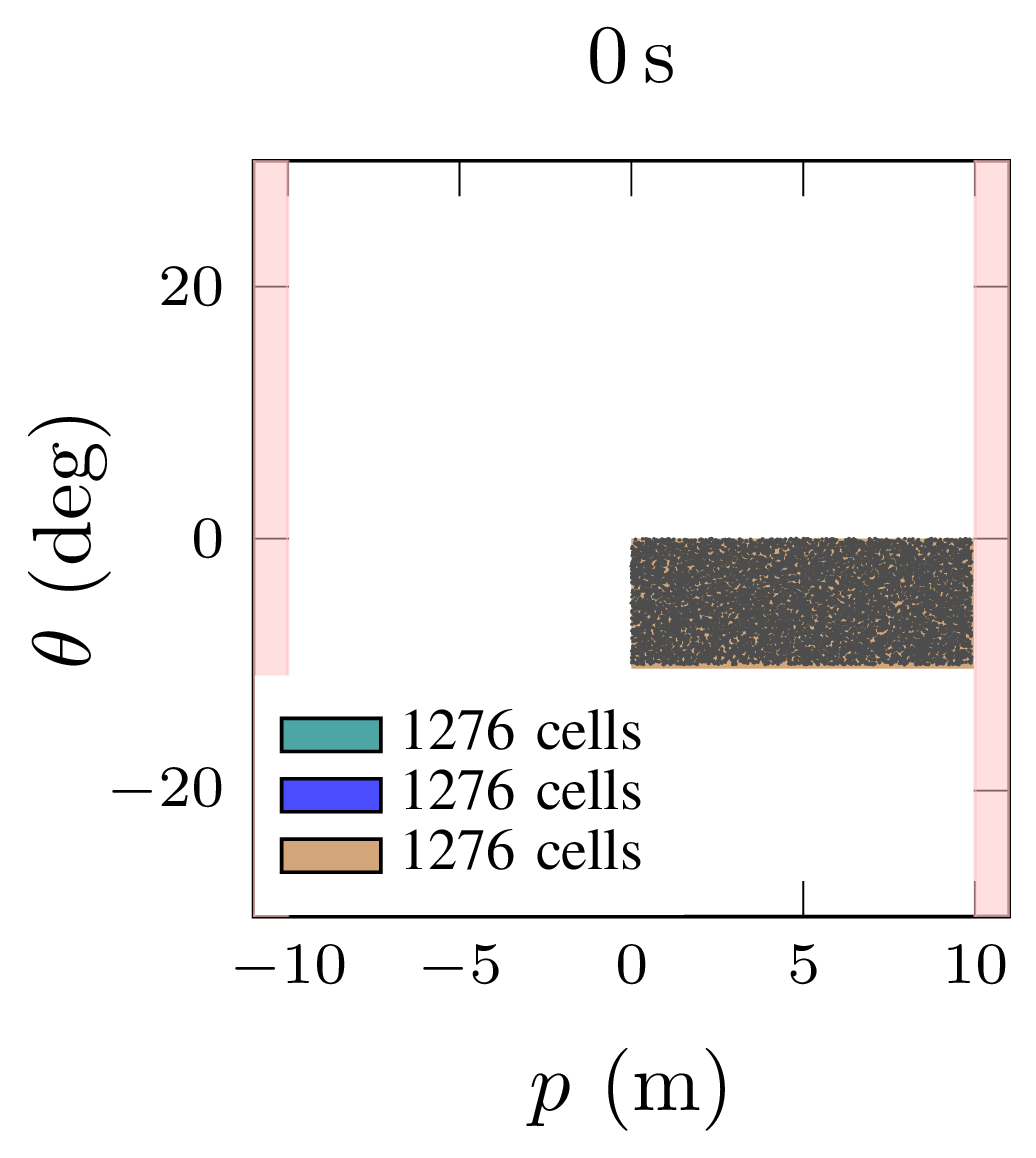}}
	\hspace{-1.0em}
	\subfloat{\includegraphics{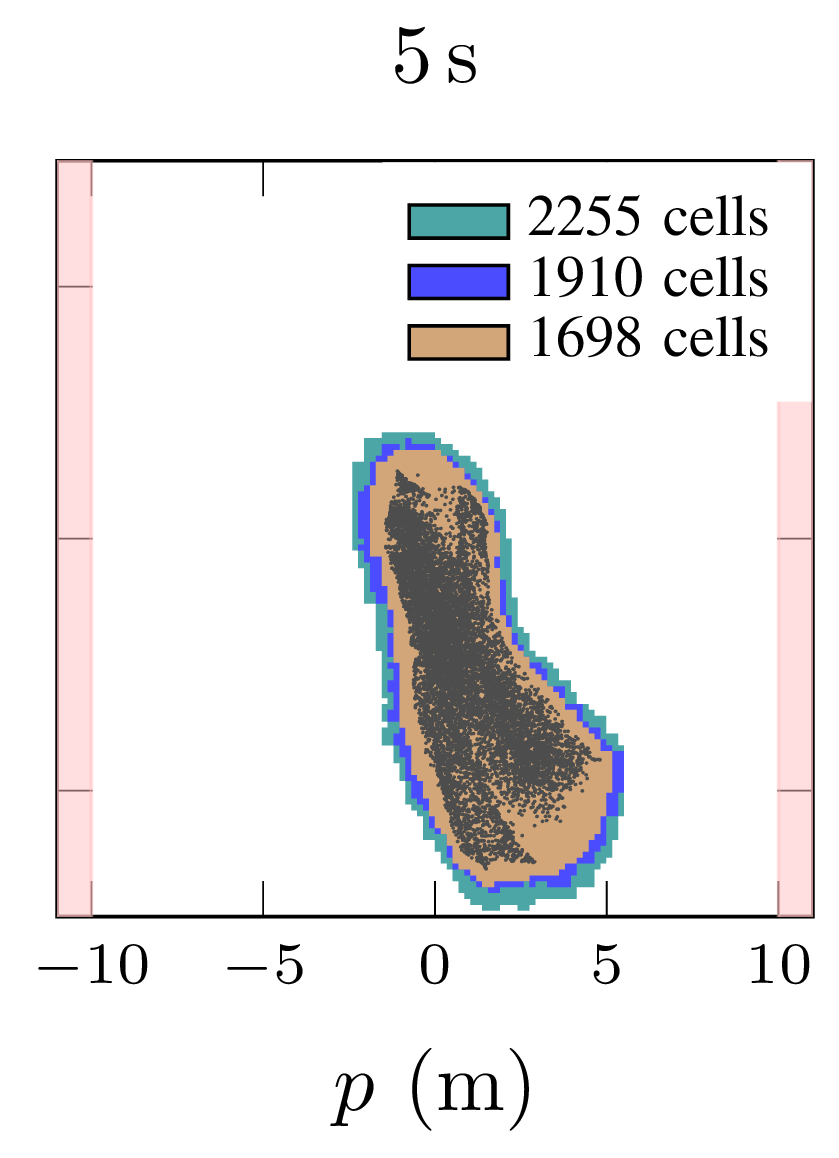}}\hspace{-1.0em}
	\subfloat{\includegraphics{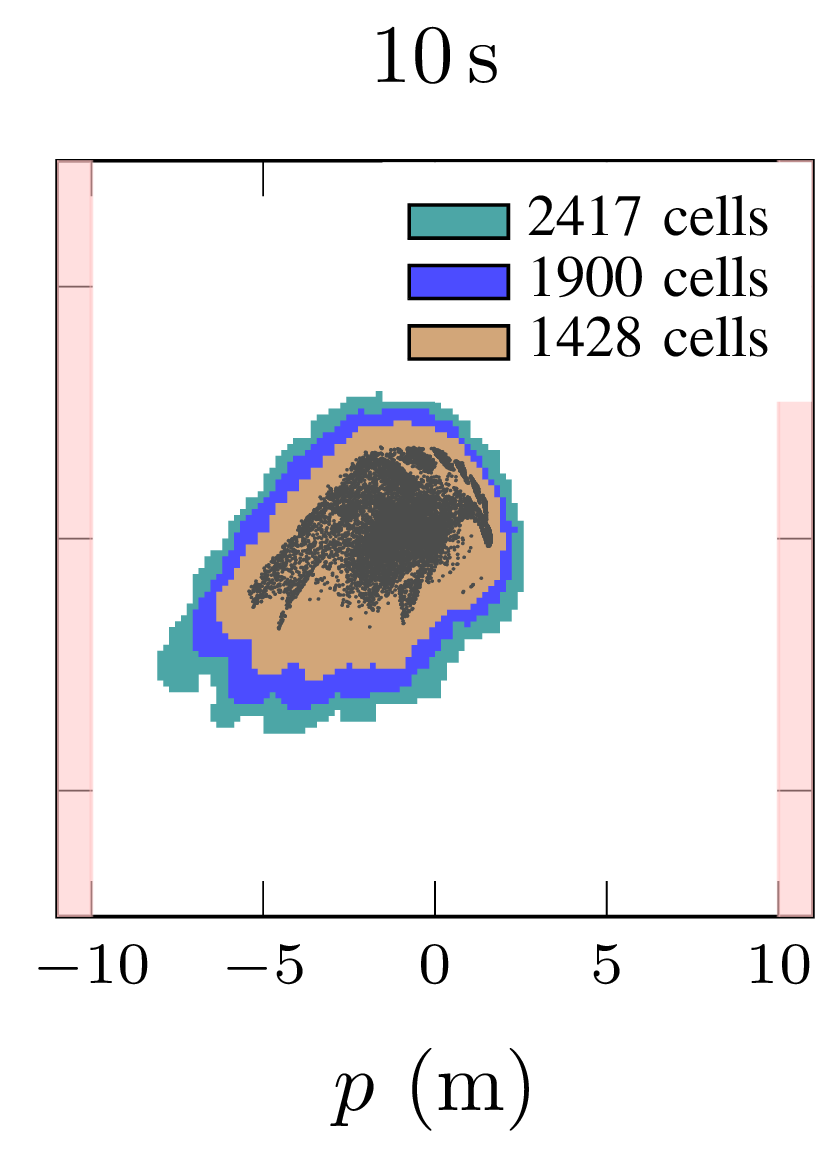}}\hspace{-1.0em}
	\subfloat{\includegraphics{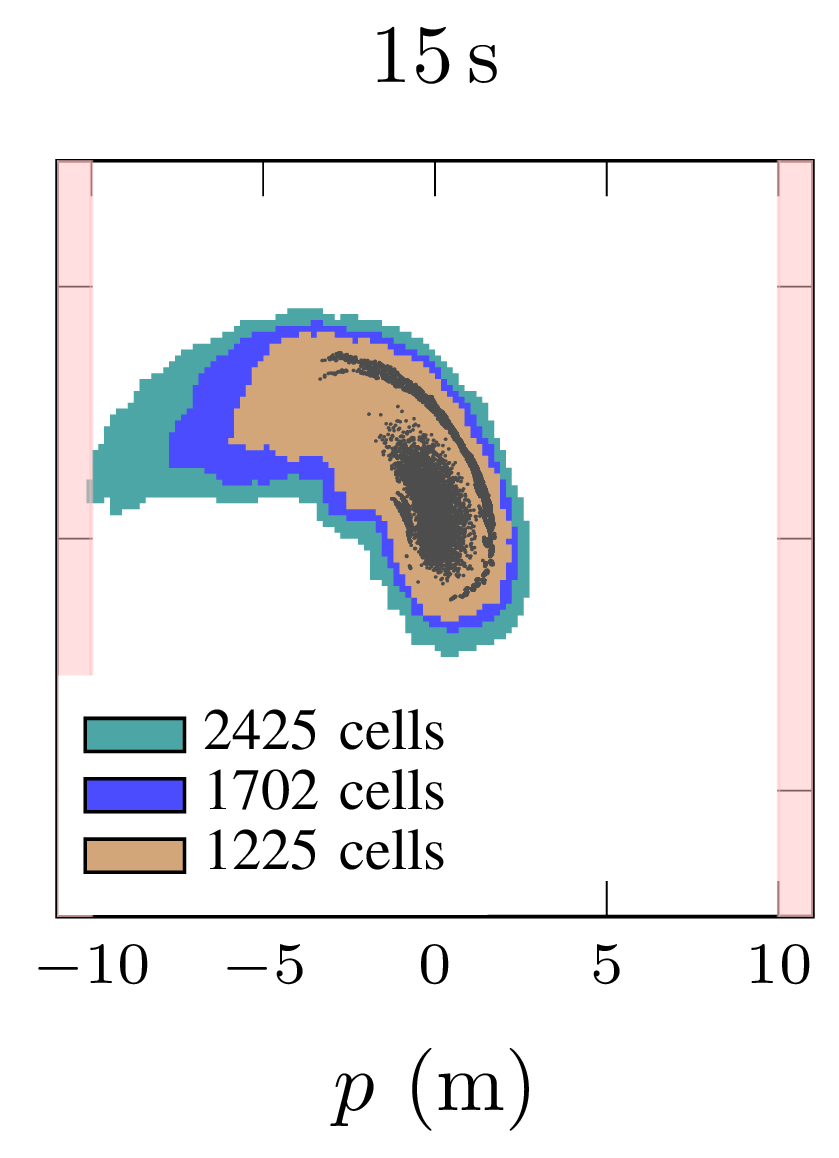}}\hspace{-1.0em}
	\subfloat{\includegraphics{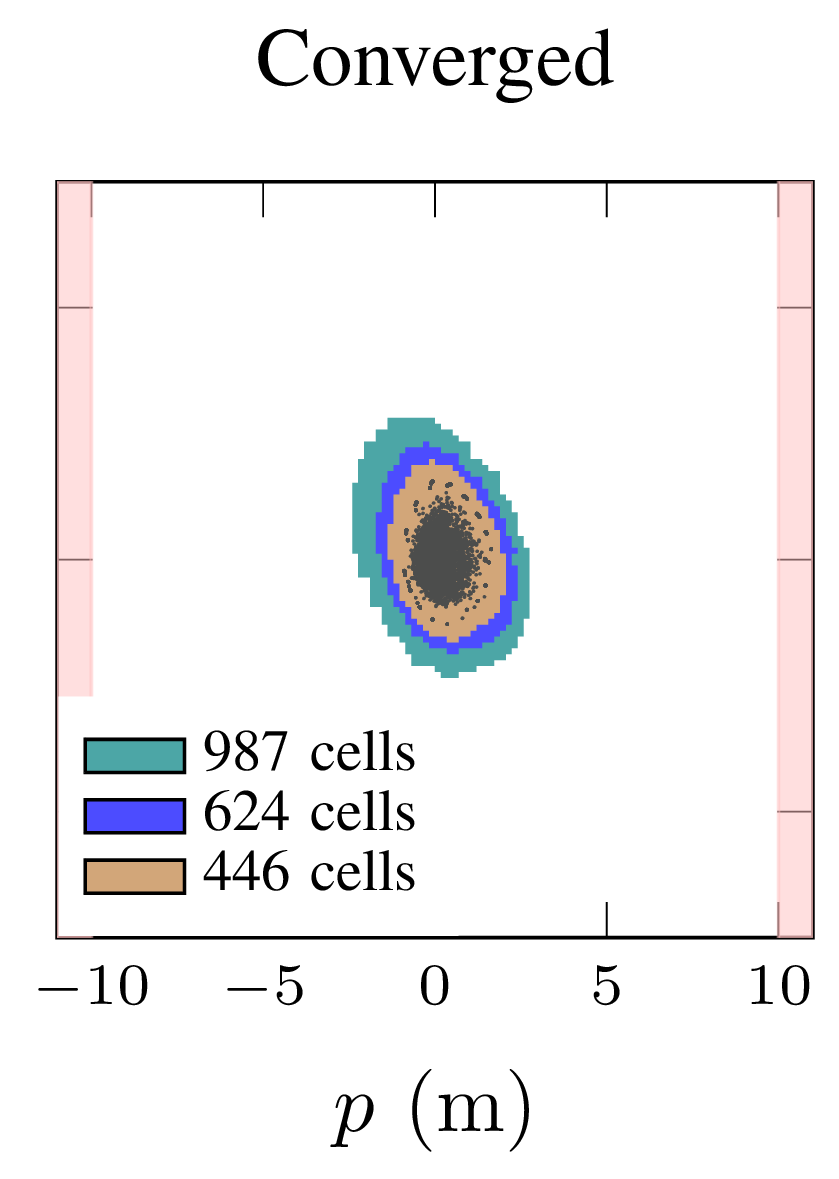}}
	\caption{Reachable sets over time using three different methods in the taxiing system.
		The colors \textcolor{black!70}{black}, \textcolor{teal!70}{teal}, 
		\textcolor{blue!70}{blue}, and \textcolor{brown!70}{brown} are used to
		represent simulations, baseline, $1$-step, and $2$-step methods, respectively.
		The color \textcolor{pink}{pink} denotes unsafe regions.
		Top: the height and width of the discretization cell are doubled, and the initial set is defined with $p\in[-\qty{9}{\meter}, \qty{9}{\meter}]$ and $\theta\in[-\qty{10}{\degree}, \qty{10}{\degree}]$. The baseline and $1$-step methods fail to prove safety of the system at $\qty{3}{\second}$, whereas $2$-step method succeeds.
		Bottom: the image-based controller is trained with less episodes, and the initial set is defined with $p\in[\qty{0}{\meter}, \qty{9.9}{\meter}]$ and $\theta\in[-\qty{10}{\degree}, \qty{0}{\degree}]$. The baseline method fails to prove safety of the system at $\qty{15}{\second}$, whereas $1$-step and $2$-step methods succeed.
	}
	\label{fig:taxi_system_additional}
\end{figure*}

The importance of our accuracy improvements becomes more apparent when parameters of the case study are altered.
To reduce verification time, we could consider a coarser grid with fewer cells.
However, this results in more abstraction error.
Another possible variant could consider a modified controller that converges more slowly.
The convergence speed of the controller is important for verification, 
as it reduces the size of the reachable set and can compensate for overapproximation error in the analysis.
%
We detail results on these variants in Fig.~\ref{fig:taxi_system_additional}.
In both scenarios, 
reachable sets computed by the baseline method extend beyond the taxiway, and 
P1 cannot be verified. 
The proposed methods effectively mitigate overapproximation, and successfully verify the safety of P1.

\section{Appendix for Advanced Emergency Braking System}
\label{appendix:aebs}
\subsection{Network Architecture and Generated Images}
We detail the network architectures for both convolutional and transformer variants in Table \ref{tab:network_architectures} and compare the real and generated images for both network variants in Fig.~\ref{fig:gan_images_aebs}.

\begin{table}[!h]
	\centering
	\normalsize
	\caption{Architectures of NNs in braking system; $\mathsf{ResBlock}$ and $\mathsf{SelfAttention}$ blocks refer to Self-Attention GAN~\cite{zhang2019self}.}
	\label{tab:network_architectures}
	\begin{subtable}[t]{0.5\textwidth}
		\centering
		\caption{Generator (convolutional network)}
		\begin{tabular}{c}
			\hhline{=}
			$z \in \mathbb{R}^4 \sim N(0, I); c \in \mathbb{R}$
			\\
			\hline
			$\mathsf{Concat} (c, z) \in \mathbb{R}^5 $
			\\
			\hline
			$\mathsf{Dense} \rightarrow 2\times2\times128$, $\mathsf{BN}$
			\\
			\hline
			$(4\times4)\times128$, $\text{stride}=2\ \mathsf{ConvTranspose}$, $\mathsf{BN}$, $\mathsf{ReLU}$ 
			\\
			\hline
			$(4\times4)\times64$, $\text{stride}=2\ \mathsf{ConvTranspose}$, $\mathsf{BN}$, $\mathsf{ReLU}$ 
			\\
			\hline
			$(4\times4)\times32$, $\text{stride}=2\ \mathsf{ConvTranspose}$, $\mathsf{BN}$, $\mathsf{ReLU}$
			\\
			\hline
			$(3\times3)\times1$, $\text{stride}=1\ \mathsf{ConvTranspose}$, $\mathsf{Tanh}$
			\\
			\hhline{=}
		\end{tabular}
	\end{subtable}
	\vspace{2mm}
	
	\begin{subtable}[t]{0.5\textwidth}
		\centering
		\caption{Controller (convolutional network)}
		\begin{tabular}{c}
			\hhline{=}
			Grayscale image $\hat{o} \in \mathbb{R}^{32\times32\times1}$
			\\
			\hline
			$(3\times3)\times16$, $\text{stride}=2,\ \text{padding}=1\ \mathsf{Conv}$, $\mathsf{ReLU}$ 
			\\
			\hline
			$(3\times3)\times32$, $\text{stride}=2,\ \text{padding}=1\ \mathsf{Conv}$, $\mathsf{BN}$, $\mathsf{ReLU}$ 
			\\
			\hline
			$(3\times3)\times64$, $\text{stride}=2,\ \text{padding}=1\ \mathsf{Conv}$, $\mathsf{BN}$, $\mathsf{ReLU}$ 
			\\
			\hline
			$(3\times3)\times128$, $\text{stride}=2,\ \text{padding}=1\ \mathsf{Conv}$, $\mathsf{BN}$, $\mathsf{ReLU}$
			\\
			\hline
			$\mathsf{Dense} \rightarrow 1\  (\hat{d} \in \mathbb{R}^1)$ 
			\\
			\hline
			$\mathsf{Concat} (\hat{d}, v) \in \mathbb{R}^2 $
			\\
			\hline
			$\mathsf{Dense} \rightarrow 400$, $\mathsf{ReLU}$
			\\
			\hline
			$\mathsf{Dense} \rightarrow 300$, $\mathsf{ReLU}$ 
			\\
			\hline
			$\mathsf{Dense} \rightarrow 1$, $\mathsf{Clamp}(0, 1)$ 
			\\
			\hhline{=}
		\end{tabular}
	\end{subtable}
\end{table}

\begin{table}[!h]
	\centering
	\normalsize
	\ContinuedFloat
	\begin{subtable}[t]{0.5\textwidth}
		\centering
		\caption{Generator (transformer network)}
		\begin{tabular}{c}
			\hhline{=}
			$z \in \mathbb{R}^4 \sim N(0, I); c \in \mathbb{R}$
			\\
			\hline
			$\mathsf{Concat} (c, z) \sim \mathbb{R}^5 $
			\\
			\hline
			$\mathsf{Dense} \rightarrow 4\times4\times256$ 
			\\
			\hline
			$\mathsf{ResBlock}$ up, $256$ 
			\\
			\hline
			$\mathsf{ResBlock}$ up, $256$ 
			\\
			\hline
			$\mathsf{SelfAttention}$, $256$ 
			\\
			\hline
			$\mathsf{ResBlock}$ up, $256$ 
			\\
			\hline
			$\mathsf{BN}$, $\mathsf{ReLU}$, $(3\times3)\times3$, $\text{stride}=1\ \mathsf{Conv}$, $\mathsf{Tanh}$
			\\
			\hhline{=}
		\end{tabular}
	\end{subtable}
	\vspace{2mm}
	
	\begin{subtable}[ht]{0.5\textwidth}
		\centering
		\caption{Controller (transformer network)}
		\begin{tabular}{c}
			\hhline{=}
			RGB image $\hat{o} \in \mathbb{R}^{32\times32\times3}$
			\\
			\hline
			$\mathsf{ResBlock}$ down, $128$ 
			\\
			\hline
			$\mathsf{SelfAttention}$, $128$  
			\\
			\hline
			$\mathsf{ResBlock}$ down, $128$ 
			\\
			\hline
			$\mathsf{ResBlock}$, $128$ 
			\\
			\hline
			$\mathsf{ResBlock}$, $256$, $\mathsf{ReLU}$
			\\
			\hline		
			$\mathsf{Dense} \rightarrow 1$, $\mathsf{Sigmoid}$,  $(\hat{d} \in \mathbb{R}^1)$ 
			\\
			\hline
			$\mathsf{Concat} (\hat{d}, v) \in \mathbb{R}^2 $
			\\
			\hline
			$\mathsf{Dense} \rightarrow 400$, $\mathsf{ReLU}$
			\\
			\hline
			$\mathsf{Dense} \rightarrow 300$, $\mathsf{ReLU}$ 
			\\
			\hline
			$\mathsf{Dense} \rightarrow 1$, $\mathsf{Clamp}(0, 1)$ 
			\\
			\hhline{=}
		\end{tabular}
	\end{subtable}
\end{table}

\begin{figure}[!h]
	\centering
	\tikzsetnextfilename{real_images_generated_images}
	\input{figures/real_images_generated_images}
	\caption{Real images and generated images for the braking system. The grayscale images are from the convolutional network; the color images are from the transformer network. }
	\label{fig:gan_images_aebs}
\end{figure}


%

\subsection{Quantitative Results of Verifying the Braking System}
Fig. 6 in the main text illustrates the set of states guaranteed to satisfy the safety property for different controllers in the braking system. Detailed quantitative results for this figure are provided in Table~\ref{table:quantitative}.

\begin{table}
	\centering
	\caption{Quantitative results of states identified as unsafe through simulations and states guaranteed to satisfy the safety property using 1-step, 2-step, and 3-step methods for different controllers in the braking system. Conv: Convolutional network variant; TF: Transformer network variant.}
	\label{table:quantitative}
	\begin{tabular}{c|c c c c}
		\hline
		& Sim (Unsafe) & $1$-step & $2$-step & $3$-step \\ 
		\hline
	Conv ($5$Hz)& 3463 & 0 & 384 & 2669 \\
	Conv ($10$Hz)& 3521 & 390 & 2892 & 5739 \\
	Conv ($20$Hz)& 3617 & 2859 & 5883 & 6035 \\
	TF ($10$Hz)& 3517 & 390 & 2890 & 5707 \\
	\hline
	\end{tabular}
\end{table}

\section{Verification Times}
\label{appendix:run_time}
We also provide the verification times for both the taxiing and braking systems in Table~\ref{tab:run_time}. 
For example, verifying the braking system with a $\qty{10}{\hertz}$ controller using $1$-step method requires approximately $125.14$ GPU-hours (around 5 days). 
This time 
increases to ${374.26}$ GPU-hours (around 15 days) for the $3$-step method.

Additionally, as the system dimension exceeds two dimensions, the verification time could grow exponentially. 
While improving the efficiency of neural network verification tools is beneficial for the efficiency of closed-loop verification, it is crucial to design a more effective verification strategy.

\begin{table}[!h]
	\centering
	\normalsize
	\caption{Verification times for all cells in the state graph. Note that these times are normalized to the times required by a single machine. These times are provided for reference, as experiments are conducted in parallel on multiple machines with different CPU and GPU configurations. The reported times are cumulative results across all machines.}
	\begin{tabular}{c|c|cc}
		\hline
		\multirow{2}{*}{} & \multirow{2}{*}{\parbox{1cm}{\centering Taxiing \\ system}} & \multicolumn{2}{c}{Braking system} \\
		\cline{3-4}
		& &Convolutional & Transformer \\
		\hline
		$1$-step method & $\qty{51.16}{\hour}$ & $\qty{125.14}{\hour}$ & $\qty{2630.88}{\hour}$\\
		
		$2$-step method & $\qty{3954.77}{\hour}$ & $\qty{168.67}{\hour}$ & $\qty{3113.45}{\hour}$\\
		
		$3$-step method & N/A & $\qty{374.26}{\hour}$ & $\qty{3609.23}{\hour}$\\
		\hline
	\end{tabular}
	\label{tab:run_time}
\end{table}

%% file: figures/real_images_generated_images_aircraft.tex
\begin{tikzpicture}[
	image/.style = {text width=0.14\textwidth, 
		inner sep=0pt, outer sep=0pt},
	node distance = 0.65mm and 0.5mm
	] 
	\node [image] (true0)
	{\includegraphics[width=\linewidth]{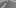}};
	\node [left=of true0, font=\small, align=center] {$p = \qty{-0.57}{\meter}$ \\ $\theta = \qty{19.40}{\degree}$};
	\node [font=\small,above=0.7ex] at (true0.north) {real};
	\node [image, right=of true0] (gen0)
	{\includegraphics[width=\linewidth]{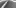}};
	\node [font=\small,above=0.3ex] at (gen0.north) {generated};
	
	\node [image, below=of true0] (true1)
	{\includegraphics[width=\linewidth]{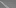}};
	\node [left=of true1, font=\small, align=center] {$p = \qty{-9.77}{\meter}$ \\ $\theta = \qty{-29.11}{\degree}$};
	\node [image, right=of true1] (gen1)
	{\includegraphics[width=\linewidth]{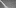}};

\end{tikzpicture}

%% file: figures/real_images_generated_images.tex
\begin{tikzpicture}[
	image/.style = {text width=0.1\textwidth, 
		inner sep=0pt, outer sep=0pt},
	node distance = 0.65mm and 0.5mm
	] 
	\node [image] (true0)
	{\includegraphics[width=\linewidth]{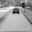}};
	\node [left=of true0, font=\small, align=center] {$\qty{7.2}{\meter}$};
	\node [font=\small,above=0.7ex] at (true0.north) {real};
	\node [image, right=of true0] (gen0)
	{\includegraphics[width=\linewidth]{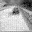}};
	\node [font=\small,above=0.3ex] at (gen0.north) {generated};
	\node [image, right=of gen0] (true1)
	{\includegraphics[width=\linewidth]{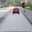}};
	\node [font=\small,above=0.7ex] at (true1.north) {real};
	\node [image, right=of true1] (gen1)
	{\includegraphics[width=\linewidth]{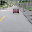}};
	\node [font=\small,above=0.3ex] at (gen1.north) {generated};

	\node [image,below=of true0] (true2)
	{\includegraphics[width=\linewidth]{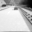}};
	\node [left=of true2, font=\small, align=center] {$\qty{15.0}{\meter}$};
	\node [image, right=of true2] (gen2)
	{\includegraphics[width=\linewidth]{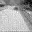}};
		\node [image, right=of gen2] (true3)
	{\includegraphics[width=\linewidth]{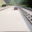}};
	\node [image, right=of true3] (gen3)
	{\includegraphics[width=\linewidth]{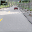}};
\end{tikzpicture}